\documentclass[11pt]{article}

\usepackage{cite}
\usepackage[utf8]{inputenc}
\usepackage{amsmath}
\usepackage{amssymb}
\usepackage{amsthm}
\usepackage[dvips]{graphicx}
\DeclareGraphicsExtensions{.eps}
\usepackage[caption=false,font=footnotesize,subrefformat=parens,labelformat=parens]{subfig}
\usepackage{array}
\usepackage{lineno}
\usepackage[linesnumbered,ruled,lined,noend]{algorithm2e}
\usepackage{fullpage}
\usepackage{setspace}
\singlespace
\usepackage{fancyhdr}
\newcommand*{\size}[1]{\left\lvert #1 \right\rvert}
\newcommand*{\dist}[1]{\left\lVert #1 \right\rVert}
\newcommand*{\var}[1]{\mathrm{var}\!\left( #1 \right)}

\newcommand*{\e}[1]{$\times 10^{#1}$}
\newcommand*{\argmin}{\operatornamewithlimits{arg\,min}}

\newcommand*{\vc}[1]{\vec{#1}}
\newcommand*{\set}[1]{#1}
\newcommand*{\sset}[1]{\mathcal{#1}}
\newcommand*{\setbrace}[1]{\left\lbrace #1 \right\rbrace}
\newcommand*{\setmean}[1]{\bar{#1}}
\newcommand*{\mi}[1]{\(#1\)}

\newcommand*{\eq}[1]{Equation~(\ref{#1})}
\SetKwInput{KwInit}{\nl initialize}
\newtheorem{theorem}{Theorem}
\newtheorem{lemma}{Lemma}

\newtheorem{corollary}{Corollary}

\title{A New Parallel Adaptive Clustering \\ and its Application to Streaming Data}
\author{McLaughlin, Benjamin R.\ S.\ and Kang, Sung Ha}
\date{\today}

\begin{document}
\maketitle

\begin{abstract}
    This paper presents a parallel adaptive clustering (PAC) algorithm to automatically classify data while simultaneously choosing a suitable number of classes.  Clustering is an important tool for data analysis and understanding in a broad set of areas including data reduction, pattern analysis, and classification.  However, the requirement to specify the number of clusters in advance and the computational burden associated with clustering large sets of data persist as challenges in clustering.  We propose a new parallel adaptive clustering (PAC) algorithm that addresses these challenges by adaptively computing the number of clusters and leveraging the power of parallel computing.  The algorithm clusters disjoint subsets of the data on parallel computation threads. We develop regularized set \mi{k}-means to efficiently cluster the results from the parallel threads.  A refinement step further improves the clusters.  The PAC algorithm offers the capability to adaptively cluster data sets which change over time by reusing the information from previous time steps to decrease computation.  We provide theoretical analysis and numerical experiments to characterize the performance of the method, validate its properties, and demonstrate the computational efficiency of the method.
    
    This work was funded by NAVSEA.  Distribution Statement A: Approved for Public Release, Distribution is Unlimited.
\end{abstract}

\section{Introduction}

Automatic classification is a task arising in various data processing applications \cite{everitt2011cluster}.  As data sets increase in size and complexity, applications demand higher degrees of efficiency and autonomy from data processing algorithms.  Clustering methods separate data into classes that exemplify and expose distinctive characteristics and trends within the data.  One strength of clustering approaches is that they discover and exploit latent relationships within the data, rather than relying on predefined class descriptors.

Clustering methods are roughly divided into three types: centroid-based, density-based, and distribution-based methods.  Centroid-based methods partition the data space into a set of \mi{k} clusters based on the distance between each datum and its associated cluster descriptor (\textit{e.g.}, centroids).  The \mi{k}-means method \cite{macqueen1967some} is a classic example of this class, as well as the closely-related \mi{k}-medians \cite{jain1988algorithms} and \mi{k}-medoids \cite{kaufman2009finding} methods.  There are also fuzzy variants of these methods \cite{jain1988algorithms} in which each datum belongs to multiple clusters with varying degree.
Density-based methods, such as the popular DBSCAN algorithm \cite{ester1996density}, partition data into clusters based on the data density.  In this approach, two data belong to the same cluster when they are connected by a densely-populated region in data space.  One feature of this method is that it finds cluster edges, enabling it to automatically compute the number of clusters and identify clusters of arbitrary shape.  However, density-based approaches can struggle to partition data sets with irregular variance.  Distribution-based methods, such as mixture models, model each cluster as a statistical distribution \cite{banfield1993model, fraley1998many}.  Each cluster can be represented as a continuous distribution function, which is advantageous for applications when it is necessary to synthesize new data samples.

The task of finding the global optimizer of the \mi{k}-means problem is \mi{NP}-hard in general.  A number of methods have been developed to efficiently compute nearly-optimal partitions \cite{lloyd1982least,hartigan1979algorithm,sparks1973algorithm} and to choose the initial cluster centroids judiciously \cite{pena1999empirical}.  Choosing a good set of initial cluster centroids can improve convergence and provide a solution that is closer to being globally optimal.  The k-means++ algorithm uses a probabilistic approach to choose initial centroids that are diverse but representative of the data set \cite{arthur2007k}.  Initializing in this manner decreases the likelihood of either assigning many centroids to a single high-density area, or of assigning many centroids to outliers.  More recently, yinyang \mi{k}-means  introduced a progressive series of filters to further improve efficiency by eliminating unnecessary computations \cite{ding2015yinyang}.

Many improvements in clustering techniques have been realized, but a persistent challenge for most centroid-based clustering methods is that the number of clusters, \mi{k}, must be chosen \textit{a priori}.  In many applications, the correct number of clusters is not known in advance.  One strategy to overcome this challenge is to repeat the clustering for many choices of \mi{k}, and then identify the best \mi{k} using a cluster quality metric such as the silhouette score \cite{rousseeuw1987silhouettes} or Dunn's index \cite{dunn1974well}.  The regularized \mi{k}-means algorithm, introduced in \cite{kang2011regularized}, presents a method for addressing this issue by computing both the number of classes and the class descriptors.

As multi-core computing architectures have become common, parallel algorithms to cluster large sets of data have received a great deal of attention.  Hierarchical and density-based clustering approaches can benefit from growing individual clusters or subsets of the hierarchy tree in parallel \cite{arlia2001experiments, olman2008parallel, olson1995parallel}.  In the \mi{k}-means setting, parallel algorithms distribute the work of computing the distances between each data and the cluster centroids (for example, \cite{stoffel1999parallel, kantabutra2000parallel, li1989parallel}).  This strategy has been implemented in numerous ways to take advantage of specific parallel architectures such as GPU computing \cite{li2013speeding}, message-passing systems \cite{zhang2013parallel}, and the map-reduce framework \cite{zhao2009parallel}.  These methods demonstrate impressive performance, but rely on the number of clusters being known in advance.  Methods for analyzing data streams have been developed for both the sequential (\textit{e.g.}, \cite{guha2003clustering,zhang1996birch}) and parallel \cite{garg2006pbirch} paradigms, but questions persist regarding whether and how the number of clusters should change as new data arrives.

In this paper we present a parallel adaptive clustering (PAC) algorithm that uses parallel computation to improve performance while applying the contribution of the regularized \mi{k}-means to automatically choose the number of clusters.  This new algorithm first forms a large set of small clusters over subsets of the data using shared memory parallelism which scales very well when the number of data is much larger than the number of processors.  An efficient algorithm is derived to reduce the parallel result into a small set of clusters over the full data set. The contributions of this paper are
\begin{itemize}
    \item a new efficient parallel adaptive clustering (PAC) method,
    \item exploration of the analytical properties of the proposed algorithm,
    \item a new algorithm for clustering data streams.
\end{itemize}
The paper begins with a review of the regularized \mi{k}-means method and a new improvement to the algorithm in Section~\ref{sec:regKmeans}.  Section~\ref{sec:algorithm} introduces the new PAC algorithm and Section \ref{sec:analysis} presents mathematical analysis of the algorithm.  Numerical experiments validate the algorithm and highlight its characteristics in Section~\ref{sec:experiments}.  Section~\ref{sec:streaming} presents the new algorithm for clustering data that arrives over time, and demonstrates the ability of the algorithm to flexibly adjust the partition as new data arrives, using far fewer computations than the naive approach.  The discussion of parallel adaptive clustering is concluded in Section~\ref{sec:conclusion}.

\section{Review of Regularized $k$-means} \label{sec:regKmeans}

One of the principal centroid-based approaches that adaptively computes the number of clusters is regularized \mi{k}-means \cite{kang2011regularized}.  The regularized \mi{k}-means method simultaneously computes both the number of clusters, \mi{k}, and the set of clusters, \mi{\sset{G} = \setbrace{\set{G}_i}_{i=1}^k}, by minimizing the global regularized \mi{k}-means energy function,
\begin{equation}  \label{eq:global_energy}
    E_{RKM}(X,\sset{G},\lambda) := \lambda\sum\limits_{i=1}^{k}\dfrac{1}{\size{\set{G}_i}} + \sum\limits_{i=1}^{k}\sum\limits_{\vc{x} \in \set{G}_i} \dist{\setmean{g}_i - \vc{x}}^{2} \,,
\end{equation}
where \mi{\vec{x} \in \set{X} \subset \mathbb{R}^d} are the data to be clustered, \mi{\setmean{g}_i \in \mathbb{R}^d} is the centroid of cluster \mi{\set{G}_i}, and \mi{\size{\set{G}_i}} is the number of data in cluster \mi{\set{G}_i}.  The regularization parameter, \mi{\lambda > 0}, is chosen to balance cluster size and cluster variance.  The set of clusters, \mi{\sset{G}}, is a partition of \mi{X}: \mi{\cup_{i=1}^{k}\set{G}_i = \set{X}} and \mi{\set{G}_i \cap \set{G}_j = \emptyset \enskip \forall i\neq j}.

The regularized \mi{k}-means algorithm initially assigns all data to a single cluster, \mi{k=1} where  \mi{\set{G}_1 = \set{X}}, and takes a greedy approach to minimize the regularized \mi{k}-means energy function.  During a single iteration, each datum is individually examined to determine whether the global energy can be reduced by reassigning the datum to a different cluster while holding all else constant.  The change in global energy resulting from moving datum, \mi{x}, currently assigned cluster \mi{\set{G}_i}, to a different cluster, \mi{\set{G}_j}, can easily be expressed as \mi{\Delta E_{RKM}(\vc{x}, \set{G}_i,\set{G}_j,\lambda)} and computed directly from the energy \eq{eq:global_energy}.
If \mi{\Delta E_{RKM}(\vc{x}, \set{G}_i,\set{G}_J,\lambda) < 0}, then \mi{\vc{x}} will be reassigned to cluster \mi{\set{G_j}}, where
\begin{equation} \label{eq:best_cluster}
    j := \argmin\limits_{J=1 \dots k+1,\enskip J \neq i} \Delta E_{RKM}(\vc{x}, \set{G}_i,\set{G}_J,\lambda)
\end{equation}
If \mi{j = k+1}, a new cluster is created.  The regularized \mi{k}-means method is presented in Algorithm \ref{alg:rkm}.

The main benefit of the regularized \mi{k}-means algorithm is that it does not require the number of clusters to be specified \textit{a priori}, which is advantageous when the best choice of \mi{k} is unknown prior to clustering.  This approach requires the selection of the regularization parameter, \mi{\lambda}, to balance the terms in the global energy.  It was shown in \cite{sandberg2010unsupervised} and \cite{kang2011regularized} that the clustering result is not sensitive to the value of \mi{\lambda}, as \mi{\lambda} may vary over a large interval without causing the number of clusters to change.  Furthermore, it was demonstrated that when the number of clusters produced by both methods is the same, the regularized \mi{k}-means algorithm can produce clusters identical to those obtained by classical \mi{k}-means.

The algorithm presented in \cite{kang2011regularized} lacks a way to remove existing clusters although this is sometimes necessary to reduce the total energy.  We present a modified regularized \mi{k}-means algorithm to address this situation by pausing after each iteration through the data set to identify existing clusters that should be merged.  Merging two clusters, \mi{G_i} and \mi{G_j}, produces a change in the global energy equal to
\begin{equation}  \label{eq:mergeEnergy}
\begin{split}
  \Delta E_{RKM}&\left(G_i,G_j,\lambda\right) = \dfrac{\size{\set{G}_i}\size{\set{G}_j}}{\size{\set{G}_i}+\size{\set{G}_j}}\dist{\setmean{g}_i - \setmean{g}_j}^2 \\ &+\lambda\left\lbrack\dfrac{1}{\size{G_i}+\size{\set{G}_j}} - \dfrac{1}{\size{\set{G}_i}} - \dfrac{1}{\size{\set{G}_j}}\right\rbrack\,.
\end{split}
\end{equation}
The clusters \mi{\set{G}_i} and \mi{\set{G}_j} should be merged when this quantity is negative.  This condition is computationally efficient to evaluate and provides a way to remove spurious clusters caused by sensitivity to data order.  This new modification is included in Algorithm~\ref{alg:rkm} (lines 11-13).


\begin{algorithm}[t]
    \SetInd{0.25em}{0.75em}
	\caption{Regularized \mi{k}-means algorithm (with modification in lines 11-14)}   	\label{alg:rkm}
	\KwIn{\mi{\set{X}};  \mi{\lambda}; \textit{ITER\_MAX}; \textit{TOL}}
	\KwInit{set \mi{k\leftarrow1}; assign all \mi{\vc{x}\in\set{X}} to \mi{\set{G}_1}}
	\Repeat{Energy change is less than \textit{TOL} or maximum number of iterations exceeds \textit{ITER\_MAX}}{
		\ForEach{\mi{\vc{x}\in\set{X}}}{
			Let \mi{\vc{x}\in\set{G}_i}; \enskip compute \mi{j =\argmin\limits_{J=1 \dots k+1} \Delta E_{RKM} \left(\vc{x},\set{G}_i,\set{G}_J,\lambda\right)}\; 
			\If{\mi{\Delta E_{RKM}\left(\vc{x},\set{G}_i,\set{G}_j,\lambda\right)<0}}{
				\eIf{\mi{j \leq k}}{
					Reassign \mi{\vc{x}} from \mi{\set{G}_i}  to \mi{\set{G}_j}; Update \mi{\setmean{g}_i}, \mi{\setmean{g}_j}\;
				}
				{ 
					Create \mi{\set{G}_{k+1}} by assigning \mi{\vc{x}} to \mi{\set{G}_{k+1}}\;
					Update \mi{\setmean{g}_i}; set \mi{\setmean{g}_{k+1} = \vc{x}}; Set \mi{k \leftarrow k+1}\;
				}
			}
		}
		\ForEach{\mi{i,j\in 1 \dots k,\enskip i \neq j}}{
			\If{\mi{\Delta E_{RKM}\left(\set{G}_i,\set{G}_j,\lambda\right) < 0}}{
				Merge \mi{\set{G}_i} and \mi{\set{G}_j}, and set \mi{k \leftarrow k-1}\;
			}
		}
		Compute total energy according to \eq{eq:global_energy}\;
	}
	\KwOut{Set of clusters, \mi{\sset{G}}}
\end{algorithm}

\section{The PAC Approach} \label{sec:algorithm}

To adaptively cluster a given data set, the PAC algorithm partitions subsets of the data, \mi{\set{X}_p}, into sets of clusters, \mi{\sset{C}_p}, \mi{p=1 \dots n}, using parallel computation to decrease computation time.  Then, these clusters are efficiently collected into groups, \mi{\sset{G} = \left\lbrace\set{G}\right\rbrace}, using a regularized set \mi{k}-means approach.  Finally, a refinement step resolves misclassified points.  The adaptive capability of the parallel step can reduce the number of misclassified points in the refinement step.  Algorithm \ref{alg:prkm} outlines the complete parallel adaptive clustering algorithm.


\begin{algorithm}[t]
    \SetInd{0.25em}{0.75em}
	\caption{Parallel Adaptive Clustering}   	\label{alg:prkm}
	\KwIn{\mi{\set{X}}; \mi{n}; \mi{\lambda_c}; \mi{\lambda_g}; \textit{ITER\_MAX}; \textit{TOL}}
	Partition \mi{\set{X}} into \mi{n} subsets, \mi{\set{X}_1 \dots \set{X}_n}\;
	\ForEach{\mi{p = 1 \dots n}, in parallel,}{
		Use regularized \mi{k}-means (Algorithm \ref{alg:rkm}) to partition \mi{\set{X}_p} into a set of clusters, \mi{\sset{C}_p}.
	}
	Collect clusters from all parallel threads, \mi{\sset{C} = \bigcup_{p=1}^{n}\sset{C}_p}\;
	Use regularized set \mi{k}-means (Algorithm \ref{alg:rkm} with \eq{eq:group_energy}) to partition \mi{\sset{C}} into global clusters, \mi{\sset{G}}\;
	Refine the clusters \mi{\sset{G} = \left\lbrace \set{G} \right\rbrace} to obtain the final cluster configuration (Algorithm \ref{alg:refinement})\;
	\KwOut{Set of clusters, \mi{\sset{G}}}
\end{algorithm}

\subsection{Parallel Clustering}	\label{subsec:divide}
The first step of the parallel algorithm distributes the data to parallel threads.  The data set is randomly partitioned into \mi{n} disjoint subsets, \mi{\set{X} = \bigcup_{p=1}^{n} \set{X}_p}, and in parallel each \mi{\set{X}_p} is partitioned into a set of clusters, \mi{\sset{C}_p}, by minimizing the regularized \mi{k}-means energy on each thread, which is \eq{eq:global_energy} for the individual thread,
\begin{equation}  \label{eq:thread_energy}
	E_{C}(\set{X}_p,\sset{C}_p,\lambda_c) := \sum\limits_{j=1}^{\size{\sset{C}_p}}\dfrac{\lambda_c}{\size{\set{C}_{p,j}}} + \sum\limits_{j=1}^{\size{\sset{C}_p}}\sum\limits_{\vc{x} \in \set{C}_{p,j}} \dist{\setmean{c}_{p,j} - \vc{x}}^{2} \,, 
\end{equation}
where \mi{\set{C}_{p,j} \in \sset{C}_p} is the \mi{j^\text{th}} cluster in the partition of \mi{\set{X}_p}.  In this work, the regularization parameter, \mi{\lambda_c}, is chosen to be the same for all threads to avoid bias, but it is possible to vary the regularization parameter for each thread.

\subsection{Grouping}  \label{subsec:unite}
Once the clusters from the parallel threads have been accumulated, groups are formed by minimizing the regularized set \mi{k}-means energy,
\begin{equation}  \label{eq:group_energy}
	E_g(\sset{C},\sset{G},\lambda_g) := \lambda_g\sum\limits_{i=1}^{\size{\sset{G}}}\dfrac{1}{\size{\set{G}_i}} + \sum\limits_{i=1}^{\size{\sset{G}}}\sum\limits_{\set{C} \in \set{G}_i} \size{\set{C}}\cdot\dist{\setmean{g}_i - \setmean{c}}^{2} \,.
\end{equation}
Here, \mi{\sset{C}} is the set of all clusters from the parallel threads: \mi{\sset{C} = \setbrace{ \set{C}_{i,j} }_{\forall i,j}} where \mi{\sset{C}_p=\setbrace{ \set{C}_{i,j} }_{i=p}}, and \mi{\lambda_g} is the regularization parameter for the grouping step.
The groups being formed are \mi{\set{G}_i \subseteq \sset{C}}, and \mi{\size{\set{G}_i} = \sum_{\set{C} \in \set{G}_i} \size{\set{C}}}.
The energy \eq{eq:group_energy} is minimized using Algorithm~\ref{alg:rkm}, where \eq{eq:group_energy} is substituted for \eq{eq:global_energy}.  This minimization must be done on a single thread, but it is computationally efficient because it does not require the examination of individual data points.
Reassigning \mi{\set{C}} from \mi{\set{G}_i} to another existing cluster \mi{\set{G}_j, j \in 1 \dots k, j \neq i} results in a change to the global energy equal to
\begin{equation} \label{eq:grouping_marginal_energy_existing}
\begin{aligned}
    \Delta E_g&(\set{C},\set{G}_i,\set{G}_j,\lambda_g) = \\ &\dfrac{\lambda_g\size{\set{C}}}{\size{\set{G}_i}(\size{\set{G}_i}-\size{\set{C}})}  - \dfrac{\lambda_g\size{\set{C}}}{\size{\set{G}_j}(\size{\set{G}_j}+\size{\set{C}})} \\ &+ \dfrac{\size{\set{G}_j}\cdot\size{\set{C}}}{\size{\set{G}_j}+\size{\set{C}}}\dist{\setmean{g}_j-\setmean{c}}^2 - \dfrac{\size{\set{G}_i}\cdot\size{\set{C}}}{\size{\set{G}_i}-\size{\set{C}}}\dist{\setmean{g}_i-\setmean{c}}^2\,,
 \end{aligned}
\end{equation}
with \mi{\Delta E_g(\set{C},\set{G}_i,\set{G}_i,\lambda_g)=0}.
The change in global energy produced by moving \mi{\set{C}} to be a new cluster, \mi{\set{G}_{k+1}}, is equal to
\begin{equation} \label{eq:grouping_marginal_energy_new}
\begin{aligned}
    \Delta E_g&(\set{C},\set{G}_i,\set{G}_{k+1},\lambda_G) = \\ &\dfrac{\lambda_g\size{\set{C}}}{\size{\set{G}_i}(\size{\set{G}_i}-\size{\set{C}})}+\dfrac{\lambda_g}{\size{\set{C}}} - \dfrac{\size{\set{G}_i}\cdot\size{\set{C}}}{\size{\set{G}_i}-\size{\set{C}}}\dist{\setmean{g}_i - \setmean{c}}^2 \,.
 \end{aligned}
\end{equation}

\subsection{Refinement}  \label{subsec:refineProc}
Since the grouping step does not alter the contents of the cluster from the parallel threads, a refinement is necessary to correct poorly classified points. If \mi{\lambda_c} and \mi{\lambda_g} are chosen appropriately, then the number of misclassified points will be small.  The refinement procedure searches for points in each cluster for which reassignment to a different cluster would reduce the global energy (all else remaining constant).  All points identified for relocation are moved simultaneously and cluster centroids are updated accordingly.  Empty clusters are removed at the end of each iteration.


\begin{algorithm}[t]
    \SetInd{0.25em}{0.75em}
	\caption{Refinement} \label{alg:refinement}
	\KwIn {Set of clusters, \mi{\sset{G}}}
	\Repeat{Energy change is less than \textit{TOL} or number of iterations exceeds \textit{ITER\_MAX}}{
		\ForEach{\mi{i\in\setbrace{ 1 \dots \size{\sset{G}}}}}{
		    \ForEach{\mi{j\in\setbrace{ 1 \dots \size{\sset{G}}}}}{
			    Obtain \mi{\gamma_{i,j}} by \eq{eq:gamma}\;
	            \ForEach{\mi{\set{C} \subseteq \set{G}_i}}{
	                Let \mi{\rho := \max\limits_{\vc{x}\in\set{C}} \dist{\bar{c}-\vc{x}}}\;
		            \If {\mi{\dist{\bar{\set{G}}_i - \bar{c}} + \rho > \gamma_{i,j}}}{
			            \ForEach{\mi{\vc{x} \in \set{C}}}{
				            \If {\mi{\dist{\bar{\set{G}}_i - \vc{x}} > \gamma_{i,j}}}{
					            Compute \mi{\Delta E_{RKM}(\vc{x},\set{G}_i,\set{G}_j,\lambda_r)}\;
					            Update current best cluster for \mi{\vc{x}}\;
				            }
			            }
		            }
		        }
		    }
		}
		Reassign each \mi{\vc{x}} to the best cluster identified for it\;
		Update centroid locations\;
		Compute total energy according to \eq{eq:global_energy}\;
	}
	\KwOut {Refined clusters}
\end{algorithm}

To improve efficiency, we reuse information computed during the parallel step to identify points which should not be reassigned.  The minimum distance from a point, \mi{\vc{x}}, to its assigned centroid, \mi{\setmean{g}_i}, which is necessary for reassigning \mi{\vc{x}} to reduce the global energy is \mi{\gamma_{i,j}} such that
\begin{equation} \label{eq:gamma}
		\begin{aligned}
		&\dfrac{\lambda_g}{\size{\set{G}_i}^2-\size{\set{G}_i}} - \dfrac{\lambda_g}{\size{\set{G}_j}^2+\size{\set{G}_j}} + \dfrac{\size{\set{G}_j}}{\size{\set{G}_j}+1}\dist{\setmean{g}_j - \setmean{g}_i}^2 \\
		&\;=\dfrac{2\size{\set{G}_j}}{\size{\set{G}_j}+1}\dist{\setmean{g}_j - \setmean{g}_i}\gamma_{i,j} + \dfrac{\size{\set{G}_i}+\size{\set{G}_j}}{(\size{\set{G}_i}-1)(\size{\set{G}_j}+1)}\gamma_{i,j}^2 \,,
		\end{aligned}
	\end{equation}
where \mi{\lambda_r} is the value of the regularization parameter used in the refinement step.  Section~\ref{sec:analysis} provides additional details on the selection of the regularization parameters \mi{\lambda_c}, \mi{\lambda_g}, and \mi{\lambda_r}.

A two-stage filter inspired by \cite{ding2015yinyang} efficiently eliminates points as candidates for reassignment (Algorithm~\ref{alg:refinement}).  The first filter attempts to eliminate entire subsets of data as candidates for reassignment, while avoiding examining individual points within the subsets.  For subsets which cannot be ruled out entirely, the second filter seeks to identify individual points which are not candidates for reassignment.  The first filter requires fewer computations per datum than the second filter, while both have a reduced computational burden compared to fully computing \mi{\Delta E} for each data point.  The value \mi{\gamma_{i,j}} is computed at the beginning of each refinement iteration, and the value \mi{\rho = \max\limits_{\vc{x}\in\set{C}} \dist{\bar{c}-\vc{x}}} is computed in the initial parallel clustering.  Section~\ref{subsec:refinement_efficiency} provides more detail on \mi{\gamma_{i,j}} and the filtering process.

\section{Analysis of the PAC Algorithm} \label{sec:analysis}

Mathematical analysis informs the implementation of the PAC algorithm and the characteristics of its results.  First, examination of the energy equations provides insight into judicious selection of the regularization parameters.  A study of the regularized set \mi{k}-means quantifies a strong relationship with the original regularized \mi{k}-means method and allows us to localize the clusters prior to the refinement step.  The refinement procedure is studied to improve computational efficiency.

\subsection{Selection of the Regularization Parameter} \label{subsec:lambda_and_cluster_stats}
From the mathematical form of the regularized \mi{k}-means energy function, the effect of the regularization parameter on the clusters produced by the method can be characterized.  The regularization parameter imposes an upper bound on the total variance permitted within each cluster.
\begin{lemma} \label{lem:initial_cluster_distance_bound}
	Let \mi{\sset{C}} be a set of clusters which is a minimizer of the regularized \mi{k}-means energy \eq{eq:thread_energy} on a particular parallel thread. Then \mi{\max\limits_{\vc{x}\in\set{C}} \dist{\setmean{c} - \vc{x}}^2 \leq {\lambda_c} \quad {\forall \enskip \set{C}\in\sset{C}}}.
\begin{proof}
	Let \mi{\set{C}\in\sset{C}} be a non-empty cluster, \mi{\size{\set{C}}>0}.  Since \mi{\sset{C}} is a minimizer of \eq{eq:thread_energy}, the marginal energy \eq{eq:grouping_marginal_energy_new} must be non-negative, implying that
	\begin{equation} \label{eq:initial_cluster_sq_distance_bound}
		\dist{\setmean{c}-\vc{x}}^2 \leq \lambda_c \left\lbrack 1 - \dfrac{1}{\size{\set{C}}} + \dfrac{1}{\size{\set{C}}^2} \right\rbrack \leq \lambda_c \quad \forall \enskip \vc{x} \in \set{C}\,.
	\end{equation}
\end{proof}
\end{lemma}
The maximum distance between any point in a cluster and the cluster centroid is bounded by the regularization parameter.  This result gives a bound on the variance of each cluster.
\begin{corollary} \label{cor:initial_cluster_variance_bound}   
	    Let \mi{\sset{C}} be a set of clusters which is a minimizer of the regularized \mi{k}-means energy \eq{eq:thread_energy}. Then \mi{{\var{\set{C}}\leq\lambda_c\enskip\forall\enskip\set{C}\in\sset{C}}}.
\begin{proof}
	Using \eq{eq:initial_cluster_sq_distance_bound} we decompose the variance of \mi{\set{C}},
	\begin{equation}
		\var{\set{C}} = \dfrac{1}{\size{ \set{C}}} \sum\limits_{\vc{x} \in  \set{C}} \dist{\setmean{c} - \vc{x}}^2 \leq \dfrac{1}{\size{ \set{C}}}\sum\limits_{\vc{x} \in  \set{C}} \lambda_c \,,
	\end{equation}
	which complete the proof.
\end{proof}
\end{corollary}

Lemma~\ref{lem:initial_cluster_distance_bound} and Corollary \ref{cor:initial_cluster_variance_bound} show that the regularization parameter governs the cluster width and bounds the variance of clusters produced by the regularized \mi{k}-means.  When regularized set \mi{k}-means is used to partition a set of clusters, \mi{\sset{C}=\left\lbrace\set{C}\right\rbrace}, analogous properties can be shown.
\begin{lemma} \label{lem:grouping_distance}
	Let \mi{\sset{G}} be a set of clusters which is a minimizer of the regularized set \mi{k}-means energy function \eq{eq:group_energy}. Then
	\begin{equation}
	    \max\limits_{\set{C}\subseteq\set{G}} \dist{\setmean{g} - \setmean{c}}^2 \leq \dfrac{\lambda_g}{\size{\set{C}}^2} \quad \forall \enskip \set{G}\in\sset{G}\,.
	\end{equation}
\begin{proof}
	Let each \mi{\set{G} \in \sset{G}} be a cluster which is partitioned into a set of non-empty clusters \mi{\set{C}\in\set{G}, \enskip \size{\set{C}}>0}. Because \mi{\sset{G}} is a minimizer of \eq{eq:group_energy}, the marginal energy function \eq{eq:grouping_marginal_energy_new} must be non-negative for \mi{\sset{G}}. Noting that \mi{\size{\set{G}} \geq \size{\set{C}} \geq 1}, this implies that for all \mi{\set{C} \subseteq \set{G} \in \sset{G}},
	\begin{equation} \label{eq:group_sq_distance_bound}
		\dist{\setmean{g} - \setmean{c}}^2 \leq \lambda_g \left\lbrack \dfrac{1}{\size{\set{C}}^2} - \dfrac{1}{\size{\set{C}}\cdot\size{\set{G}}} + \dfrac{1}{\size{\set{G}}^2}\right\rbrack \leq \dfrac{\lambda_g}{\size{\set{C}}^2} \,.
	\end{equation}
\end{proof}
\end{lemma}
\begin{lemma} \label{lem:grouping_variance}
	Let \mi{\sset{G}} be a set of clusters which minimizes \eq{eq:group_energy}.  Then the variance of \mi{\set{G}\in\sset{G}} is bounded,
	\begin{equation}  \label{eq:group_variance_bound}
		\var{\set{G}} \leq \dfrac{1}{\size{\set{G}}}\sum\limits_{\set{C} \subseteq \set{G}}\left\lbrack \dfrac{\lambda_g}{\size{\set{C}}} +  \size{\set{C}}\cdot\var{\set{C}}\right\rbrack \quad \forall \enskip \set{G} \in \sset{G} \,.
	\end{equation}
\begin{proof}
    \mi{\set{G}} is decomposed into disjoint subsets, the clusters \mi{\left\lbrace\set{C}\right\rbrace_{\set{C}\subseteq\set{G}}}.  The variance of \mi{\set{G}} can then be written,
	\begin{equation} \label{eq:group_variance_decomp}
		\var{\set{G}} = \dfrac{1}{\size{\set{G}}}\sum\limits_{\set{C} \subseteq \set{G}} \size{\set{C}}\left\lbrack\dist{\setmean{g}-\setmean{c}}^2 + \var{\set{C}}\right\rbrack  \,.
	\end{equation}
	Substituting \eq{eq:group_sq_distance_bound} into \eq{eq:group_variance_decomp} completes the proof.
\end{proof}
\end{lemma}

\begin{theorem} \label{thm:divide_and_unite_variance}
	Let \mi{\set{G} \in \sset{G}} be a global cluster formed by the parallel and grouping steps of the PAC algorithm.  Then \mi{\set{G}} has variance satisfying
	\begin{equation} \label{eq:divide_and_unite_variance}
		\var{\set{G}} \leq \lambda_c + \dfrac{\lambda_g}{\size{\set{G}}}\sum\limits_{\set{C} \subseteq \set{G}} \dfrac{1}{\size{\set{C}}} \,.
	\end{equation}
\begin{proof}
	Lemma~\ref{lem:grouping_variance} gives an upper bound on \mi{\var{\set{G}}} which depends on the variance of all \mi{\set{C} \subseteq \set{G}}.  In the parallel adaptive algorithm, each \mi{\set{C}} is a cluster formed by the regularized \mi{k}-means method which by Corollary~\ref{cor:initial_cluster_variance_bound} has variance bounded by \mi{\lambda_c}.  Substituting this result into Lemma~\ref{lem:grouping_variance} completes the proof.
\end{proof}
\end{theorem}

Observe the relationship between \eq{eq:divide_and_unite_variance} and the result of Corollary \ref{cor:initial_cluster_variance_bound}.  The variance of a global cluster, \mi{\set{G}}, depends on the variance of each \mi{\set{C}\in\set{G}} and the variance in the centroids of all \mi{\set{C}\in\set{G}}, which in turn are bounded by \mi{\frac{\lambda_g}{\size{\set{C}}^2}} and \mi{\lambda_c}, respectively.

\subsection{Relation Between Grouping and Global Minimum} \label{subsec:groupingAnalysis}
The objective function minimized by the grouping step is a generalization of the global energy function, and this relationship gives insight into the selection of regularization parameters.  In general a minimizer of \eq{eq:group_energy} is not a minimizer of \eq{eq:global_energy} because \mi{\var{\set{C}} \neq 0}.  Theorem~\ref{thm:grouping_consistency} shows that \mi{\sset{G}} is a minimizer of the regularized \mi{k}-means energy function \eq{eq:global_energy} subject to the constraint that the contents of each \mi{\set{C}\in\sset{C}} do not change.
\begin{theorem} \label{thm:grouping_consistency}
	Let \mi{\var{C}=0 \enskip} for all \mi{\set{C} \in \sset{C}} and \mi{\lambda = \lambda_g}.  Then the grouping energy \eq{eq:group_energy} is equivalent to the global energy, \eq{eq:global_energy}.
\begin{proof}
	For \mi{\lambda=\lambda_g}, the regularization term in \eq{eq:group_energy} is identical to that in \eq{eq:global_energy}.  The variance of \mi{\set{G}_i = \bigcup\limits_{\set{C} \subseteq \set{G}_i} \set{C}}, with \mi{\bigcap\limits_{\set{C} \subseteq \set{G}_i} \set{C} = \emptyset}, satisfies the property,
	\begin{equation} \label{eq:variance_decomposition}
		\var{\set{G}_i} = \dfrac{1}{\size{\set{G}_i}} \sum\limits_{\set{C} \subseteq \set{G}_i} \size{\set{C}} \left\lbrack\dist{\setmean{g}_i - \setmean{c}}^2 + \var{\set{C}} \right\rbrack \,.
	\end{equation}
	Enforcing \mi{\var{\set{C}}=0 \enskip \forall \set{C} \subseteq \set{G}_i} and substituting the result into \eq{eq:global_energy} completes the proof.
\end{proof}
\end{theorem}
In the special case that \mi{\size{\set{C}}=1}, \eq{eq:group_energy} is identical to the marginal energy from \cite{kang2011regularized} and is a generalization of regularized \mi{k}-means clustering for data with arbitrary mass.  The computation of marginal energy associated with merging two groups in \eq{eq:mergeEnergy} is still applicable in this setting.  The grouping algorithm is very efficient when \mi{\size{\sset{C}} \ll \size{\set{X}}} because the computation involves only the aggregate mean and size of each \mi{\set{C}\in\sset{C}}.

Deciding whether \mi{\set{C}} should be assigned to \mi{\set{G}} for a particular distance \mi{\dist{\setmean{g}-\setmean{c}}} depends on the sizes of \mi{\set{C}} and \mi{\set{G}} as well as on \mi{\lambda_g}.  In general, the clusters \mi{\set{C}} and \mi{\set{G}} contain many elements.  The grouping step requires \mi{\lambda_{g} \gg \lambda_{c}} due to the quantity of elements in each \mi{\set{C}}, as shown in the following theorem.
\begin{theorem}  \label{thm:lambda_and_cluster_size}
	Let \mi{\set{C}_1,\set{C}_2\in \sset{C}} be two non-empty clusters.  These two clusters can only be assigned to the same global cluster, \mi{\set{G}\in\sset{G}}, if
	\begin{equation} \label{eq:grouping_lambda_requirement}
		\dist{\setmean{g}-\setmean{c}}^2 \leq \lambda_g\left\lbrack\dfrac{1}{\size{\set{C}}^2} + \dfrac{1}{\size{\set{G}}\cdot\size{\set{C}}} + \dfrac{1}{\size{\set{G}}^2}\right\rbrack\,.
	\end{equation}
\begin{proof}
    Let \mi{\set{C}_1} be contained within any \mi{\set{G}_i \in \sset{G}} that also contains \mi{\set{C}_2}.  \mi{\set{C}_1} will be removed from \mi{\set{G}_i} to form a new cluster, \mi{\set{G}_{k+1}}, unless there exists some \mi{\set{G}_j} with \mi{1 \leq j \leq k} such that
	\begin{equation}
		 \Delta E_G (\set{C}_1,\set{G}_i,\set{G},\lambda_g) \leq  \Delta E_G (\set{C}_1,G_i,G_{k+1},\lambda_g) \,.
	\end{equation}
	From \eq{eq:grouping_marginal_energy_existing} and \eq{eq:grouping_marginal_energy_new}, this condition can be written,
	\begin{equation}
		\dfrac{\size{\set{G}}\cdot\size{\set{C}_1}}{\size{\set{G}}+\size{\set{C}_1}}\dist{\setmean{g}-\setmean{c}_1}^2 \leq \left\lbrack\dfrac{\lambda_g\size{\set{C}_1}}{\size{\set{G}}(\size{\set{G}}+\size{\set{C}_1})}+\dfrac{\lambda_g}{\size{\set{C}_1}}\right\rbrack \,,
	\end{equation}
	and \mi{\set{C}_1} will be moved into \mi{\set{G}_{k+1}} unless \eq{eq:grouping_lambda_requirement} is satisfied for some \mi{\set{G} \in \left\lbrace\set{G}_j\right\rbrace_{j=1}^{k}}.
\end{proof}
\end{theorem}
As \mi{\size{\set{G}}} and \mi{\size{\sset{C}}} increase, the impact of the regularization parameter on the resulting clusters decreases.  Because the clusters in \mi{\sset{C}} are smaller overall, the \mi{\frac{1}{\size{\set{C}}^2}} term in \eq{eq:grouping_lambda_requirement} will have the largest influence on the significance of \mi{\lambda_g}.  In order to retain balance between the terms in the energy equation, \mi{\lambda_g} must be increased quadratically as \mi{\size{\sset{C}}} increases.

\subsection{Stability of \mi{k} after Grouping} \label{subsec:stability_of_k}
\begin{figure}[htpb]
	\begin{center}
		\includegraphics[width=0.45\textwidth]{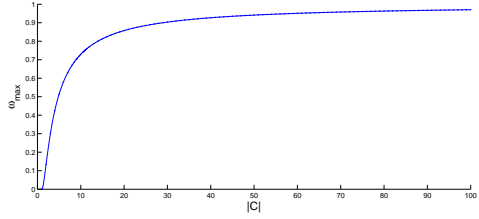}
		\caption{The quantity \mi{\lambda_g\geq\frac{\lambda_c}{\omega_{\text{max}}}} is a function of \mi{\size{C}}.  Larger \mi{\size{C}} increase the stability of \mi{k} after the grouping step, as \mi{\omega_{\text{max}}} asymptotically approaches 1.}
		\label{fig:omegaMax}
	\end{center}
\end{figure}

If the parameters are appropriately chosen then the parallel step and the grouping step should produce a set of clusters which are nearly optimal in terms of the global energy.  The number of points which must be reassigned to obtain an optimal solution should be small, which is necessary for efficiency since the refinement treats each candidate point individually.  For simplicity in parameter selection, we choose \mi{\lambda_r=\lambda_g}.  Recalling from Section~\ref{subsec:groupingAnalysis} that we require \mi{\lambda_g > \lambda_c > 0}, we introduce the parameter \mi{\omega \in (0,1)} such that \mi{\lambda_c = \omega\lambda_g}.
Judicious selection of \mi{\lambda_g} can guarantee that new clusters should not be formed during the refinement process.
\begin{theorem}  \label{thm:kStability}
	Let \mi{\lambda_c = \omega \lambda_g} for some \mi{0 < \omega < 1}.  Then for any \mi{\vc{x} \in \set{C} \subseteq \set{G} \in \sset{G}}, moving \mi{\vc{x}} to start a new cluster will cause an increase in the global energy if
	\begin{equation} \label{eq:omegaBound}
		\left(1-\omega\right)\size{\set{C}} - 2\sqrt{\omega} \geq 1 \,.
	\end{equation}
\begin{proof}
	In order for \mi{\vc{x} \in \set{G}} to form a new cluster it must be true that \mi{\size{\set{C}}\geq 1} and \mi{\size{\set{G}} \geq \max \setbrace{ \size{\set{C}}, 2 }}, therefore,
		\begin{equation}
		1 - \dfrac{1}{\size{\set{G}}} + \dfrac{1}{\size{\set{G}}^2} \geq 1 - \dfrac{1}{\size{\set{C}}} + \dfrac{1}{\size{\set{C}}^2} \,.
	\end{equation}
    The triangle inequality guarantees that for any \mi{\vc{x} \in \set{C} \subseteq \set{G}},
    \begin{equation}
        \dist{\setmean{g}-\vc{x}} \leq \dist{\setmean{g}-\setmean{c}} + \dist{\setmean{c} - \vc{x}} \,.
    \end{equation}
    Using the bounds from Lemma~\ref{lem:initial_cluster_distance_bound}, Lemma~\ref{lem:grouping_distance} and substituting the relationship \mi{\lambda_c = \omega\lambda_g},
	\begin{equation}
		\dist{\setmean{g} - \vc{x}}^2 \leq \lambda_g\left\lbrack\dfrac{1}{\size{\set{C}}} + \sqrt{\omega}\right\rbrack^2 \quad \forall \; \vc{x} \in \set{C} \subseteq \set{G} \,.
	\end{equation}
	From \eq{eq:grouping_marginal_energy_new}, it is known that reassigning \mi{\vc{x} \in \set{G}} to form a new cluster will increase the energy if
	\begin{equation}
		\lambda_g \left\lbrack 1 - \dfrac{1}{\size{\set{G}}} + \dfrac{1}{\size{\set{G}}^2}\right\rbrack \geq \dist{\setmean{g} - \vc{x}}^2 \,.
	\end{equation}
	This condition must be true if
	\begin{equation}
		\lambda_g\left\lbrack 1 - \dfrac{1}{\size{\set{C}}} + \dfrac{1}{\size{\set{C}}^2}\right\rbrack \geq \lambda_g\left\lbrack\dfrac{1}{\size{\set{C}}} + \sqrt{\omega}\right\rbrack^2 \,,
	\end{equation}
	and rearranging this inequality gives the result \eq{eq:omegaBound}.
\end{proof}
\end{theorem}
Choosing \mi{\lambda_g} sufficiently large relative to \mi{\lambda_c} prevents new clusters from forming during refinement.  If the clusters from the parallel threads are large, then \mi{\lambda_g} may be chosen close to \mi{\lambda_c} and still provide this guarantee, as illustrated in Figure~\ref{fig:omegaMax}.  In typical cases (\mi{\size{C} \gg 1}), a \mi{\lambda_g} which is large enough to affect the clustering result (as a consequence of Theorem~\ref{thm:lambda_and_cluster_size}) is also large enough to satisfy the condition of Theorem~\ref{thm:kStability}.

\subsection{Refinement Efficiency and \mi{\gamma_{i,j}}} \label{subsec:refinement_efficiency}
If the parameters for the initial clustering and grouping steps are suitably chosen, then only a small portion of the data is reassigned during the refinement.  The algorithm must determine whether each point should be reassigned.  To increase efficiency, we develop a two-tiered filtering process to quickly identify many points which should not be reassigned.  Theorem~\ref{thm:gamma} describes how the distances between centroids can be used to quickly find points which should not be reassigned.
\begin{theorem} \label{thm:gamma}
	Let \mi{\set{G}_i}, \mi{\set{G}_j}, \mi{i \neq j}, be clusters for which there exists a value \mi{\gamma_{i,j}>0} satisfying \eq{eq:gamma}.
	If \mi{\dist{\setmean{g}_i-\vc{x}}\leq\gamma_{i,j}} for a datum \mi{\vc{x}\in\set{G}}, then moving \mi{\vc{x}\in\set{G}_i} to \mi{\set{G}_j} will not decrease the total energy.
\begin{proof}
The triangle inequality guarantees that for any \mi{x \in \set{G}_i},
\begin{equation}  \label{eq:triangleInequality1}
	\dist{\setmean{g}_j - x} \geq \dist{\setmean{g}_i - \setmean{g}_j} - \dist{\setmean{g}_i-\vc{x}} \,.
\end{equation}
Substituting \eq{eq:triangleInequality1} into  \eq{eq:grouping_marginal_energy_existing}, where \mi{\set{C}=\vc{x}},
\begin{equation} \label{eq:energy_bound}
		\begin{aligned}
			\Delta& E_{RKM}(\vc{x},\set{G}_i,\set{G}_j,\lambda_g) \geq \dfrac{\size{\set{G}_j}}{\size{\set{G}_j}+1}\dist{\setmean{g}_j - \setmean{g}_i}^2 \\ &+\lambda_g\left\lbrack\dfrac{1}{\size{\set{G}_i}^2-\size{\set{G}_i}} - \dfrac{1}{\size{\set{G}_j}(\size{\set{G}_j}+1)}\right\rbrack \\
			&-\dfrac{2\size{\set{G}_j}}{\size{\set{G}_j}+1}\dist{\bar{G}_j - \bar{G}_i}\dist{\bar{G}_i - \vc{x}} \\
			&+ \left\lbrack\dfrac{\size{\set{G}_j}}{\size{\set{G}_j}+1}-\dfrac{\size{\set{G}_i}}{\size{\set{G}_i}-1}\right\rbrack\dist{\setmean{g}_i - \vc{x}}^2 \,.
		\end{aligned}
\end{equation}
\eq{eq:energy_bound} gives a lower bound on the change in energy created by moving \mi{\vc{x}} from \mi{\set{G}_i} to \mi{\set{G}_j}.  Replacing \mi{\dist{\setmean{g}_i - \vc{x}}} with \mi{\gamma_{i.j}} we obtain the left-hand side expression of \eq{eq:gamma}, which is monotone decreasing for \mi{\gamma_{i,j}>0}.  Therefore, there exists at most one \mi{\gamma_{i,j}>0} for which \eq{eq:gamma} is satisfied.  If this \mi{\gamma_{i,j}} exists, then it is guaranteed that \mi{\Delta E_{RKM}(\vc{x},\set{G}_i,\set{G}_j,\lambda_g) \geq 0} for any \mi{\vc{x} \in \set{G}_i} satisfying \mi{\dist{\setmean{g}_i - x} < \gamma_{i,j}}, and this \mi{\vc{x}} will not be moved from \mi{\set{G}_i} to \mi{\set{G}_j}.
\end{proof}
\end{theorem}

\begin{corollary} \label{cor:gamma}
	Let \mi{\set{C} \subseteq \set{G}_i}, with \mi{\rho := \max\limits_{\vc{x} \in \set{C}} \dist{\setmean{c} - \vc{x}}}.  If there exists \mi{\gamma_{i,j} > 0} which satisfies \eq{eq:gamma} and \mi{\rho \leq \gamma_{i,j} - \dist{\setmean{g}_i-\setmean{c}}}, then moving any \mi{\vc{x} \in \set{C}} from \mi{\set{G}_i} to \mi{\set{G}_j} will not decrease the total energy.
\begin{proof}
This property follows directly from the definition of \mi{\rho}, the derivation of \mi{\gamma_{i,j}}, and the triangle inequality: \mi{\forall \vc{x}\in\set{C}},
\begin{equation}   \label{eq:triangleInequality2}
	\dist{\setmean{g}_i - \vc{x}} \leq \dist{\setmean{g}_i - \setmean{c}} + \dist{\setmean{c} - \vc{x}} \leq \dist{\setmean{g}_i - \setmean{c}} + \rho \,.
\end{equation}
\end{proof}
\end{corollary}
Corollary~\ref{cor:gamma} gives a method for identifying entire clusters of points which should not be reassigned.  This forms the first layer of the refinement filter, and Theorem~\ref{thm:gamma} provides the second.  A filter using Lemma~\ref{lem:initial_cluster_distance_bound} and Lemma~\ref{lem:grouping_distance} to identify any group, \mi{\set{G}_j}, which is far enough from \mi{\set{G_i}} that no data should be reassigned from \mi{\set{G}_i} to \mi{\set{G}_j} could be implemented, but in the experiments shown here this would not add considerable savings over the filter based on Corollary~\ref{cor:gamma}.

\section{Numerical Experiments} \label{sec:experiments}

This section demonstrates the properties of the PAC method through various numerical experiments.  In these experiments, the parameter value is set to be \mi{\lambda_r = \lambda_g}, which is sufficient to guarantee that the number of clusters will not increase during the refinement (Theorem~\ref{thm:kStability}).  Due to the consequence of Theorem~\ref{thm:lambda_and_cluster_size}, \mi{\lambda_g} is chosen based on the sizes of the clusters,
\begin{equation}  \label{eq:epsilon_and_lambda}
     \lambda_g = \epsilon\left(\dfrac{\size{\set{X}}}{\size{\sset{C}}}\right)^2 \,,
\end{equation}
where it is necessary to specify the constant, \mi{\epsilon}, rather than to select \mi{\lambda_g} directly.

\subsection{Illustration of PAC Algorithm}

\begin{figure*}[htpb]
    \begin{center}
        \subfloat[\label{fig:2d_poorsep_data}]{
        	\centering
            \includegraphics[width=0.3\textwidth]{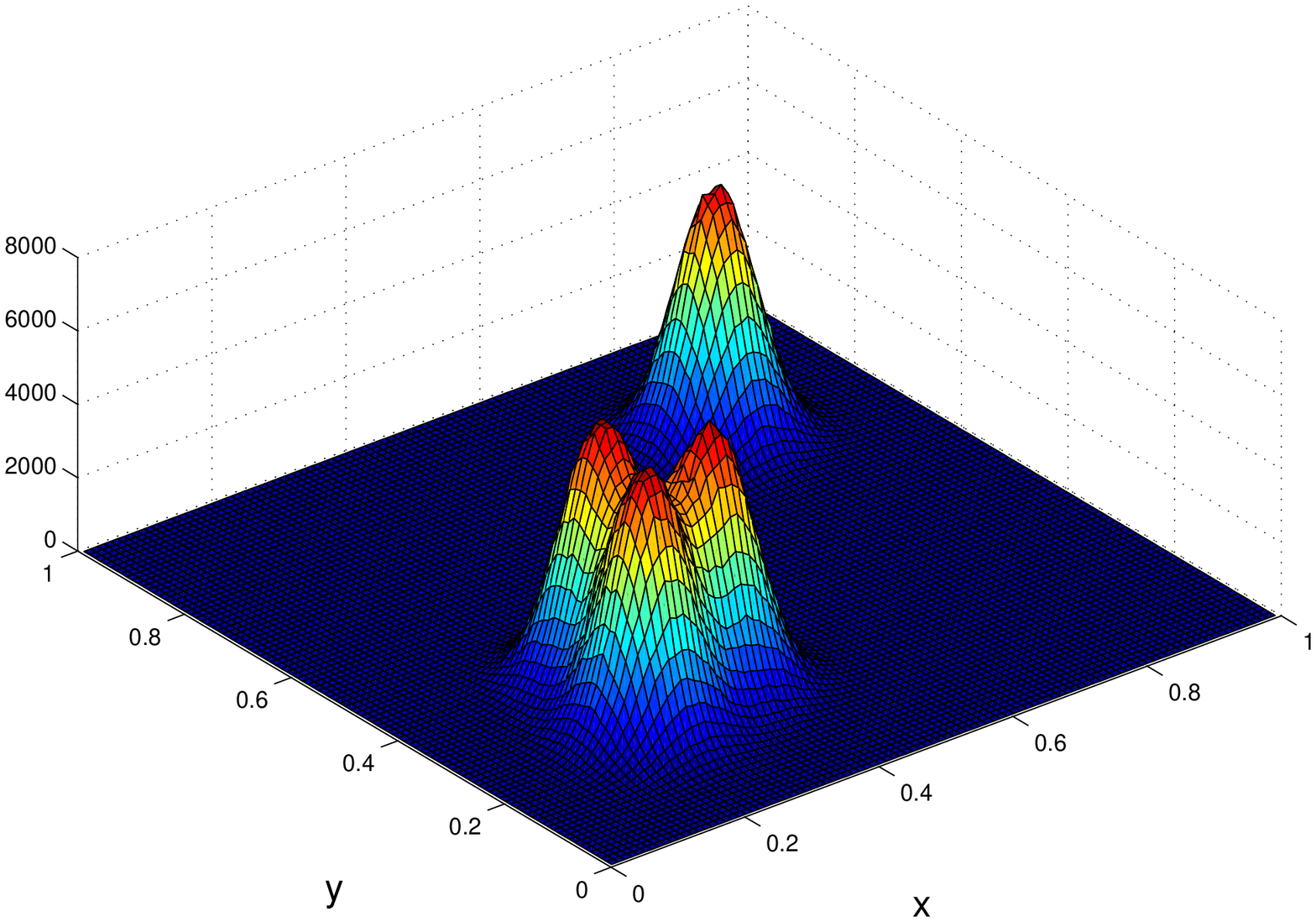}
        }
        \subfloat[\label{fig:2d_poorsep_threads}]{
            \centering
            \includegraphics[width=0.2\textwidth]{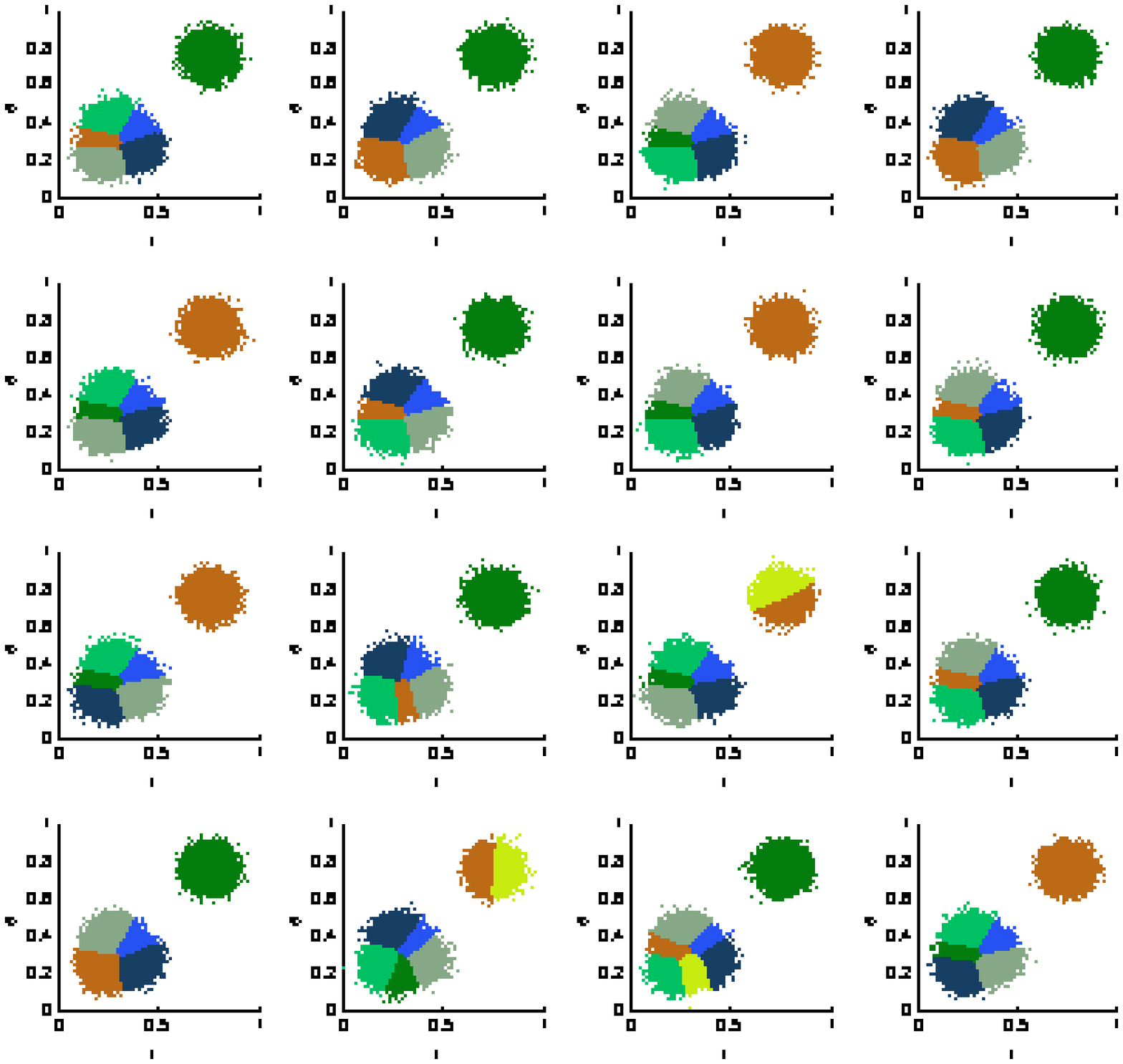}
        }
        \subfloat[\label{fig:2d_poorsep_group}]{
            \centering
            \includegraphics[width=0.2\textwidth]{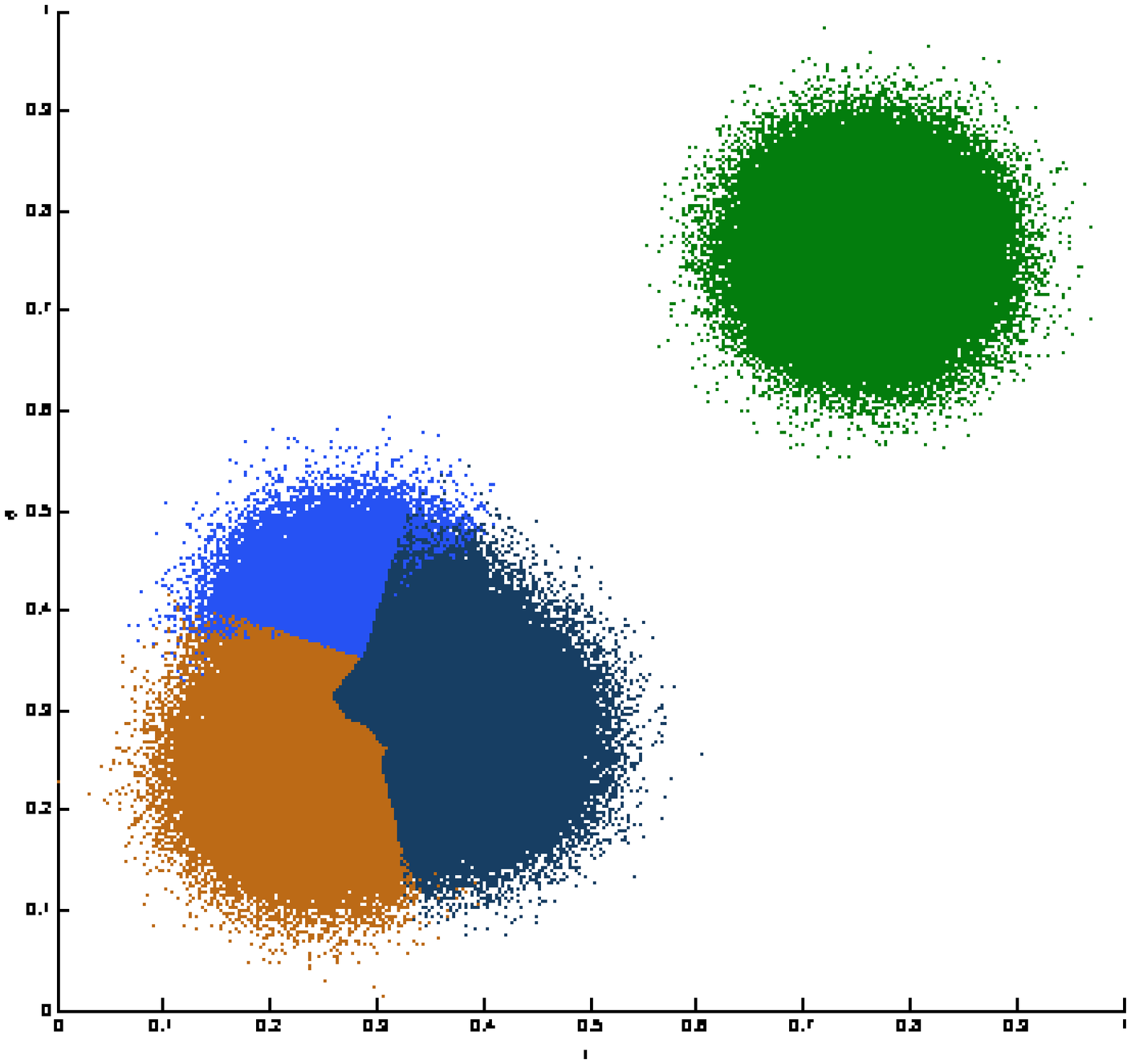}
        }
        \subfloat[\label{fig:2d_poorsep_refine}]{
            \centering
            \includegraphics[width=0.2\textwidth]{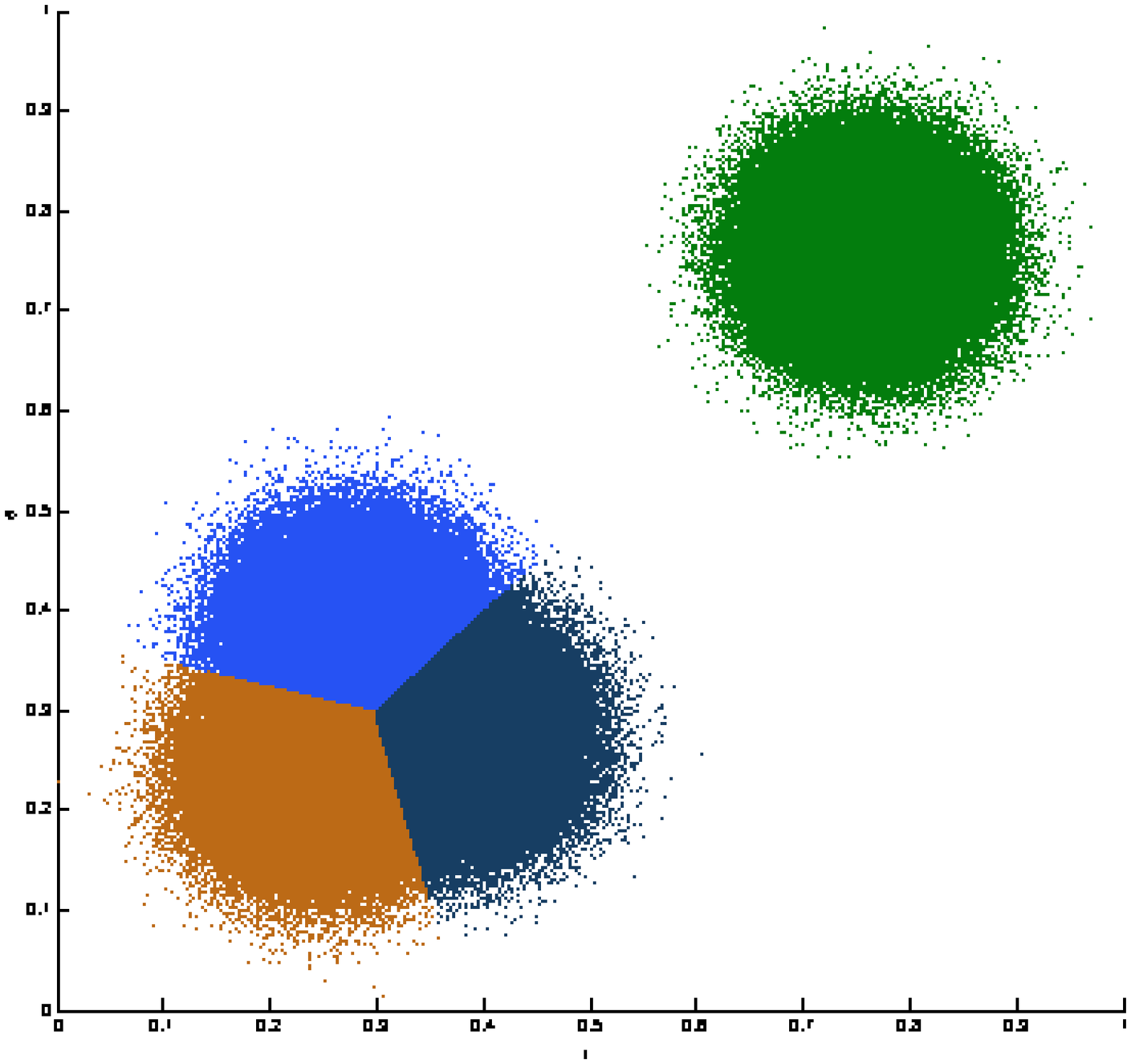}
        }
        \caption{\protect\subref*{fig:2d_poorsep_data} A set of data in \mi{\mathbb{R}^2}.   \protect\subref*{fig:2d_poorsep_threads} The result of clustering disjoint subsets of the data on 16 parallel threads (with \mi{\lambda_c = 0.05}).  Each forms a reasonable set of clusters for its subset of the data.  \protect\subref*{fig:2d_poorsep_group} The grouping step (with \mi{\epsilon = 0.05}) combines the parallel subsets into a reasonable number of global clusters, but these are overlapping due to the independence of each thread.  \protect\subref*{fig:2d_poorsep_refine} The refinement step (with \mi{\epsilon = 0.05}) sharpens the cluster boundaries.}
        \label{fig:2d_poorsep}
    \end{center}
\end{figure*}

The data set represented by the 2-d histogram in Figure~\subref*{fig:2d_poorsep_data} is a mixture of data drawn from four distinct Gaussian distributions. The means of three of those distributions lie in close proximity to one another and the tails of these distributions overlap.  This data may appear to be optimally clustered by either \mi{k=2} or \mi{k=4}.  We apply the PAC algorithm with \mi{\lambda_c = \epsilon = 0.05}
\begin{enumerate}
    \item In the parallel step, the data is randomly partitioned into 16 subsets of equal size, which are clustered on parallel computation threads using regularized \mi{k}-means (Algorithm \ref{alg:rkm}), producing the clusters in Figure~\subref*{fig:2d_poorsep_threads}.
    \item The grouping step aggregates the clusters from all parallel threads, organizing them into groups and producing the four clusters shown in Figure~\subref*{fig:2d_poorsep_group}.  These clusters are disjoint subsets of the data set, but they are overlapping in the 2-D data space.
    \item The refinement step reassigns points near the cluster boundaries (Figure~\subref*{fig:2d_poorsep_refine}).  The refinement procedure affects the cluster boundaries to a large extent while  not changing the cluster centroids as significantly.  This shows the efficiency of the refinement procedure, as only a small portion of each cluster's points are reassigned.
\end{enumerate}

\subsection{The Influence of \mi{\lambda_c}} \label{subsec:threeRingData}
\begin{figure}[htpb]
  \begin{center}
      \centering
      \includegraphics[width=0.4\textwidth]{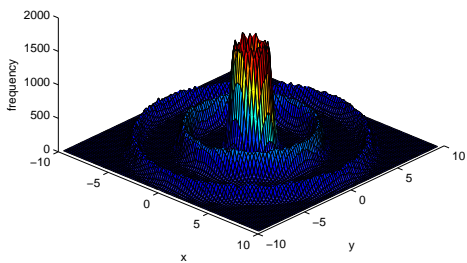}
      \caption{Set of data lying along concentric rings in Cartesian space}
      \label{fig:ring_data}
  \end{center}
\end{figure}

\begin{figure*}[htpb]
    \begin{center}
        \subfloat[\mi{\lambda_c = 1},  clusters per thread\mi{\gg 3}\label{fig:3ring_overseg}]{
            \centering
            \includegraphics[width=0.22\textwidth]{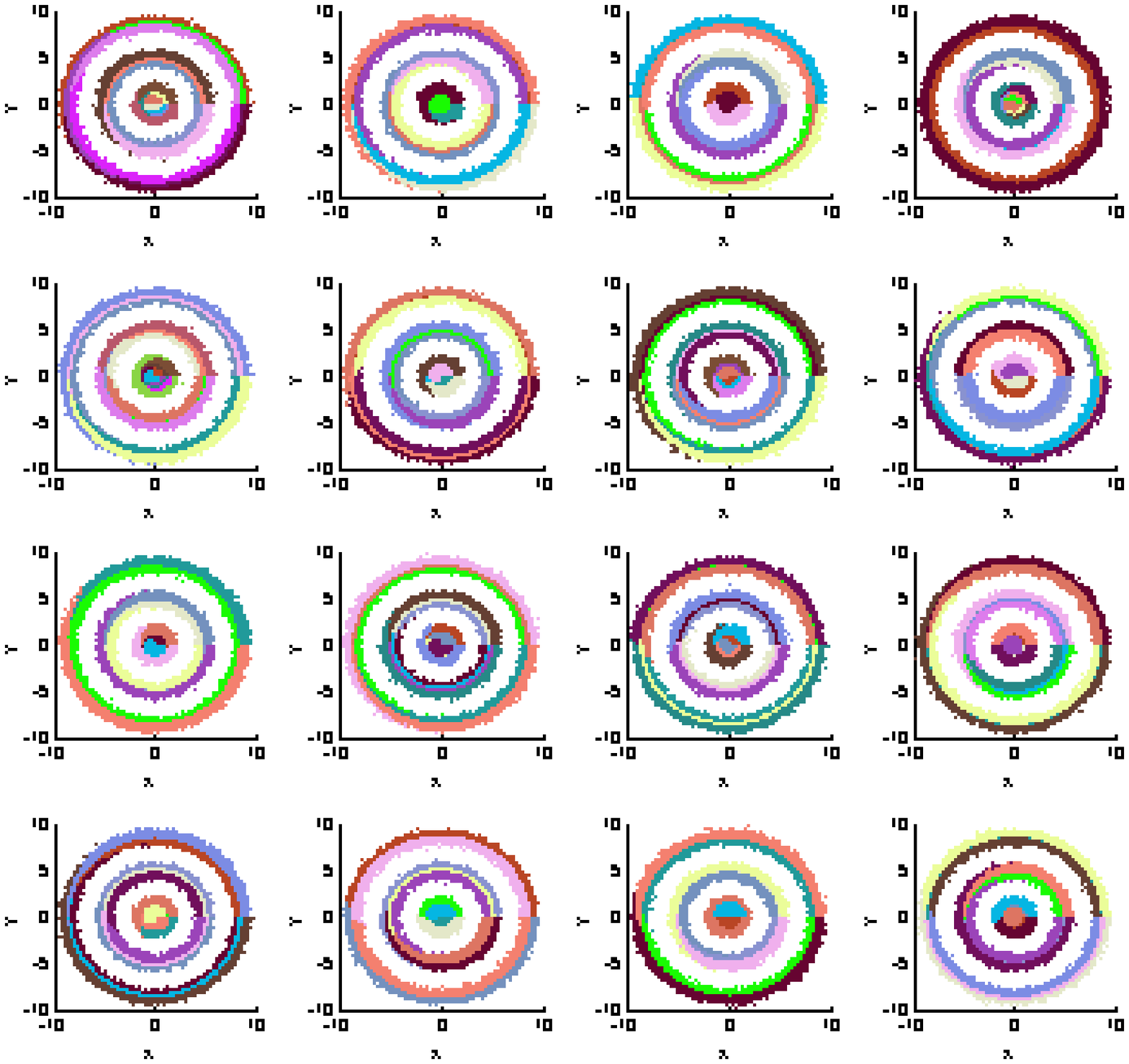}
        } \quad
        \subfloat[\mi{\lambda_c = 2},  clusters per thread\mi{\approx 3}\label{fig:3ring_normalseg}]{
            \centering
            \includegraphics[width=0.22\textwidth]{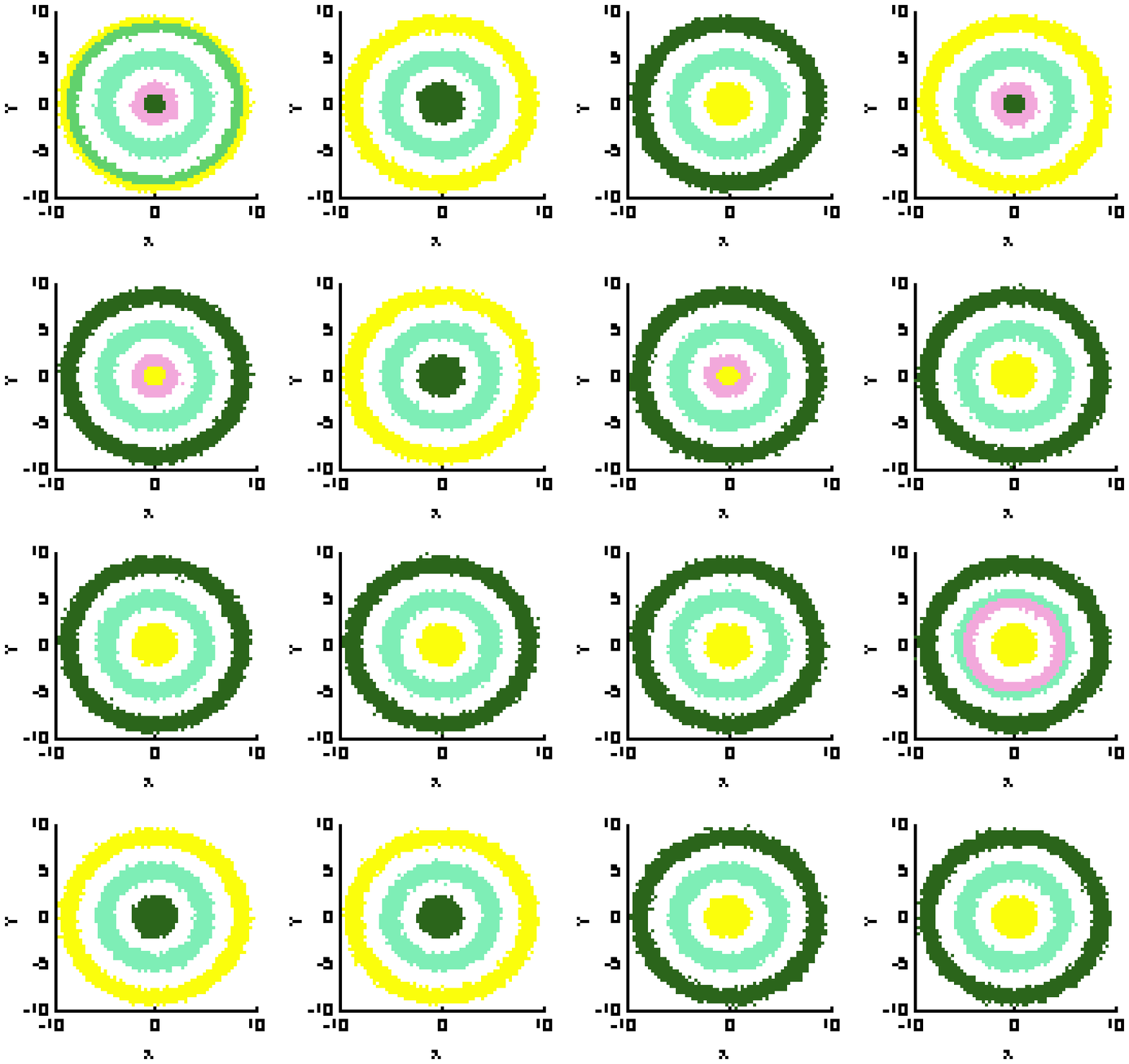}
        } \quad
        \subfloat[\mi{\lambda_c = 24},  clusters per thread\mi{<3}\label{fig:3ring_underseg}]{
            \centering
            \includegraphics[width=0.22\textwidth]{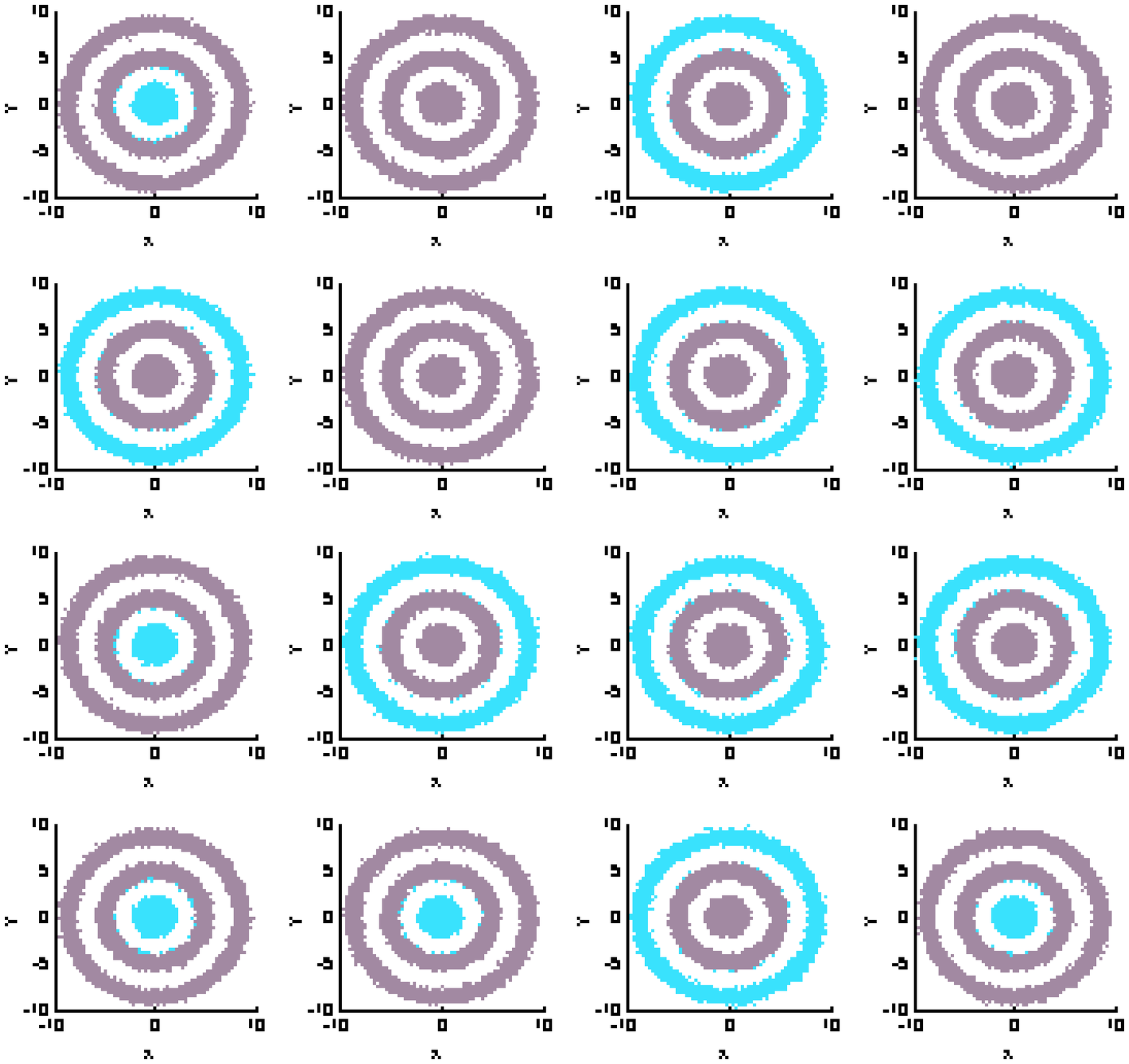}
        } \quad
        \subfloat[3-cluster final partition\label{fig:ring_final}]{
            \centering
            \includegraphics[width=0.22\textwidth]{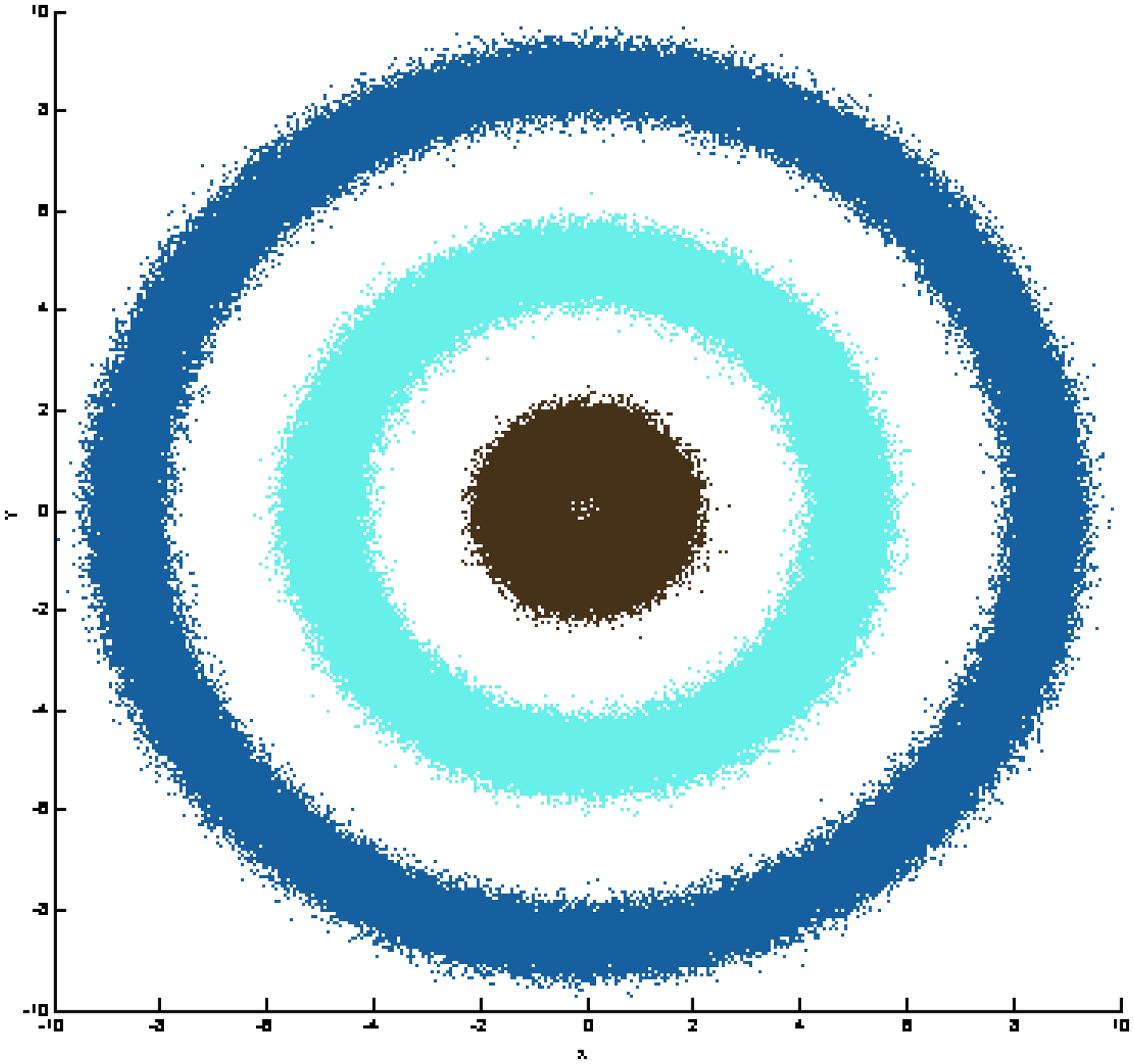}
        }
    \end{center}
    \caption{Results of using PAC to cluster (in the \mi{(r,\theta)} plane) data which lies in three concentric rings in the Cartesian plane.  In the parallel step \protect\subref*{fig:3ring_overseg}, \mi{\lambda_c=2} produces approximately three clusters per thread, while \mi{\lambda_c=1} produces far more clusters \protect\subref*{fig:3ring_normalseg} and \mi{\lambda_c=24} produces fewer \protect\subref*{fig:3ring_underseg}.  In all three cases, it is possible to choose a value \mi{\lambda_g} such that the final output of the algorithm is the three-cluster solution, \protect\subref*{fig:ring_final}.  Table \ref{tab:2d_polar_segmentation} compares the computational time to produce the three-cluster solution for each choice of \mi{\lambda_c}}.
    \label{fig:3ring_scatter}
\end{figure*}

The value of \mi{\lambda_c} affects the number of clusters produced by each thread during the parallel step, which in turn affects the subsequent steps.  Changing the values of the regularization parameters affects the efficiency of the algorithm even though the final cluster configuration may be identical.  As an example, consider a set of data lying on three concentric rings in the Cartesian plane (Figure~\ref{fig:ring_data}).  There are many choices of \mi{\lambda_c} and \mi{\epsilon} for which the parallel adaptive clustering algorithm produces the three cluster solution in Figure~\subref*{fig:ring_final}.

Due to the random assignment of data to each parallel thread, the distribution of data for each thread tends to follow the distribution of the full data set.  In Figure~\subref*{fig:3ring_normalseg}, the choice of \mi{\lambda_c = 2} results in a clustering where the number of clusters (and cluster centroids) computed by each thread is the same as or slightly larger than the number of clusters in the final configuration.  In contrast, Figure~\subref*{fig:3ring_overseg} represents an ``over-segmentation,''  where each thread produces many more clusters than will appear in the final configuration (in this case, \mi{\lambda_c = 1}).  Conversely, ``under-segmentation'' can occur when \mi{\lambda_c} is large (here, \mi{\lambda_c = 24}) and the parallel threads produce fewer clusters than will be represented in the final partition. (Figure~\subref*{fig:3ring_underseg}).


\begin{table*}[t]
	\caption{Comparison of over-segmentation and under-segmentation during parallel computation using 16 threads.}
	\label{tab:2d_polar_segmentation}
	\begin{center}
		\begin{tabular}{| r | r | r | r | r | r | r | r | r |}
		        \hline
		        \multicolumn{2}{|c|}{Parameters} & \multicolumn{2}{c|}{Parallel} & \multicolumn{2}{c|}{Grouping} & \multicolumn{2}{c|}{Refinement} & \multicolumn{1}{c|}{Total} \\
				\multicolumn{1}{|c|}{\mi{\lambda_c}} &
				\multicolumn{1}{c|}{\mi{\lambda_g}} & \multicolumn{1}{c|}{Clusters} & \multicolumn{1}{c|}{Time (s)} & \multicolumn{1}{c|}{Clusters} & \multicolumn{1}{c|}{Time (s)} & \multicolumn{1}{c|}{Clusters} & \multicolumn{1}{c|}{Time (s)} & \multicolumn{1}{c|}{Time (s)} \\[0.1em] \hline
				 1
				& 8.39\e{6} & 259 & 2.50 & 3 & 0.015 & 3 & 0.11 & 2.63 \\[0.1em]
				 2
				& 1.93\e{8} & 54 & 0.10 & 3 & 0.003 & 3 & 0.11 & 0.21 \\[0.1em]
				 24 
				& 3.04\e{6} & 29 & 0.05 & 5 & 0.005 & 3 & 1.30 & 1.36 \\[0.1em]  \hline
		\end{tabular}
	\end{center}
\end{table*}

For any of these values of \mi{\lambda_c}, one can choose \mi{\lambda_g} such that the final cluster configuration after refinement is identical to Figure~\subref*{fig:ring_final}.  The choice of \mi{\lambda_c}, however, affects the computational burden of finding that solution.  Table~\ref{tab:2d_polar_segmentation} lists some choices of \mi{\lambda_g} which produce the 3-cluster solution, as well as the associated computational time.  When the parallel threads over-segment the data, the 3-cluster solution can be achieved but requires more computation in the grouping step, to combine many small clusters into groups.  However, when the parallel threads under-segment the data, achieving the 3-cluster solution requires much more computation in the refinement step, to reassign a larger number of individual data points.  For \mi{\lambda_c=1} and \mi{\lambda_c=2} the refinement step converges in a single iteration while requiring several iterations for \mi{\lambda_c=24}.  This is investigated further in Section~\ref{subsec:refinementConvergence}.

One of the advantages of using regularized \mi{k}-means to perform the initial parallel clustering is that individual threads are not constrained by a predetermined number of clusters.  This is important for efficiency because it permits the number of clusters in each \mi{\sset{C}_p} to be optimized for \mi{\set{X}_p} which helps to prevent over- and under-segmentation.
The risk of either under-segmenting or over-segmenting the data is higher if a non-adaptive method (\textit{i.e.}, classical \mi{k}-means) is used on each parallel thread.

\subsection{Convergence of the Refinement Step}  \label{subsec:refinementConvergence}

\begin{figure}[htpb]
    \begin{center}
        \subfloat[\label{fig:2d_poorsep_convergence_percpts}]{
            \centering
            \includegraphics[width=0.23\textwidth]{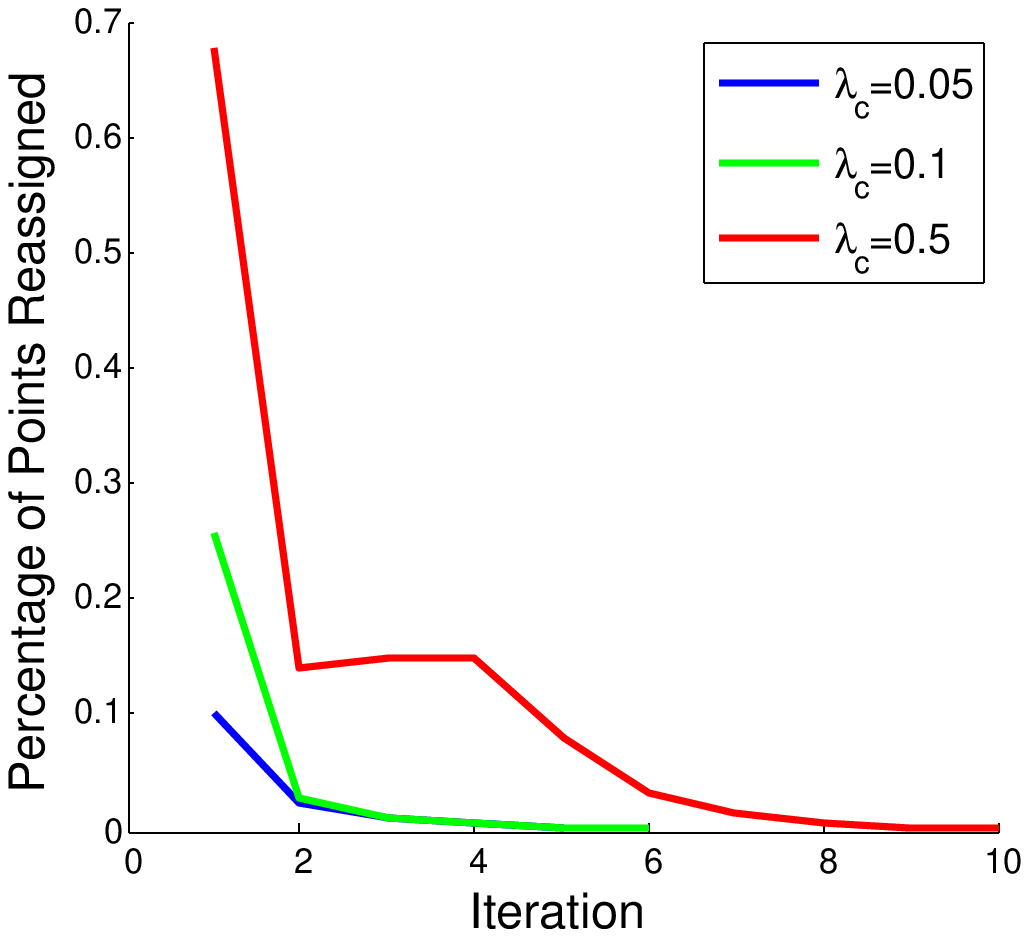}
        }
        \subfloat[\label{fig:2d_poorsep_convergence_energy}]{
            \centering
            \includegraphics[width=0.23\textwidth]{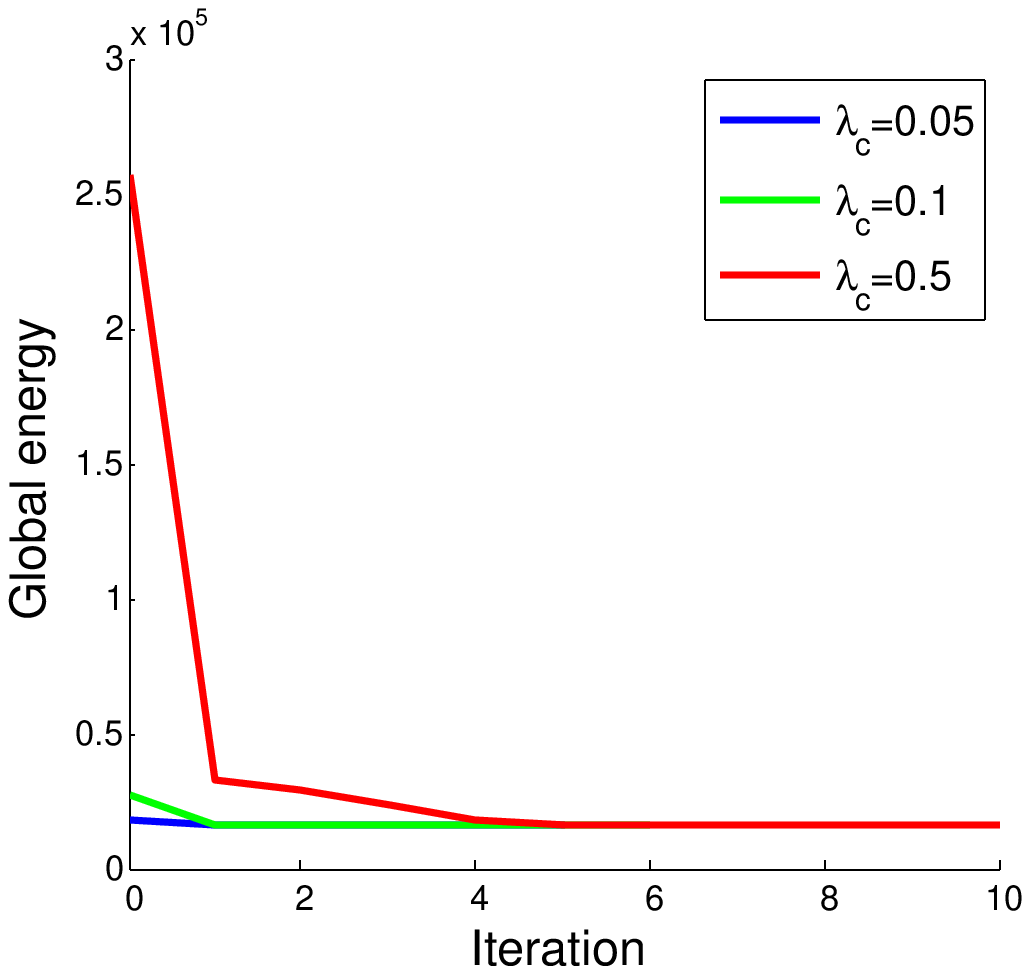}
        }
    \end{center}
    \caption{The effect of several choices of \mi{\lambda_c} on the refinement step to cluster the data set from Figure~\ref{fig:2d_poorsep}.  The percentage of the data reassigned \protect\subref*{fig:2d_poorsep_convergence_percpts} and the global energy relative to the initial energy \protect\subref*{fig:2d_poorsep_convergence_energy} are computed at each iteration.}
    \label{fig:2d_poorsep_refine_converge}
\end{figure}

\begin{figure*}[htpb]
    \begin{center}
        \subfloat[Iteration 0: After Grouping, Before Refinement\label{fig:ring_underseg_groups}]{
            \centering
            \includegraphics[height=1.5cm,width=8.7cm]{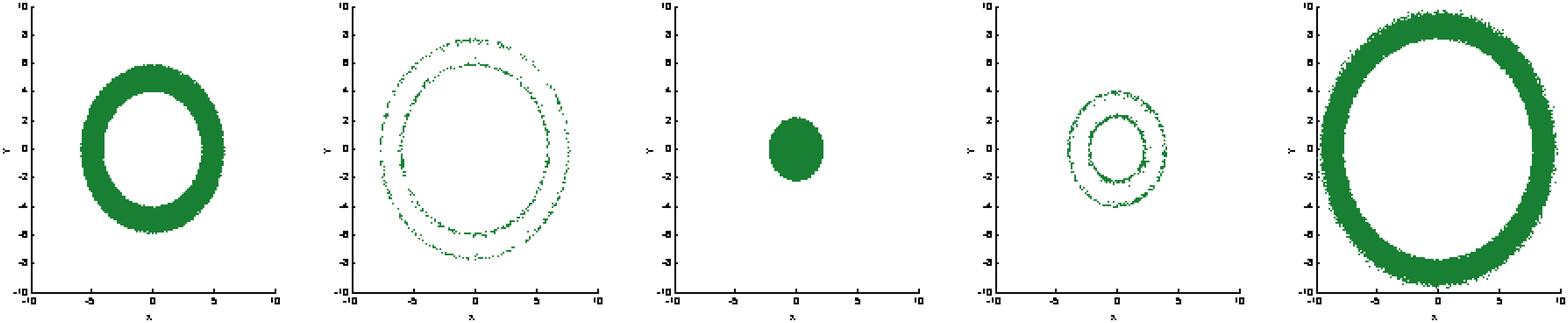}
        } \qquad
        \subfloat[Iteration 1]{
            \centering
            \includegraphics[height=1.5cm,width=7.3cm]{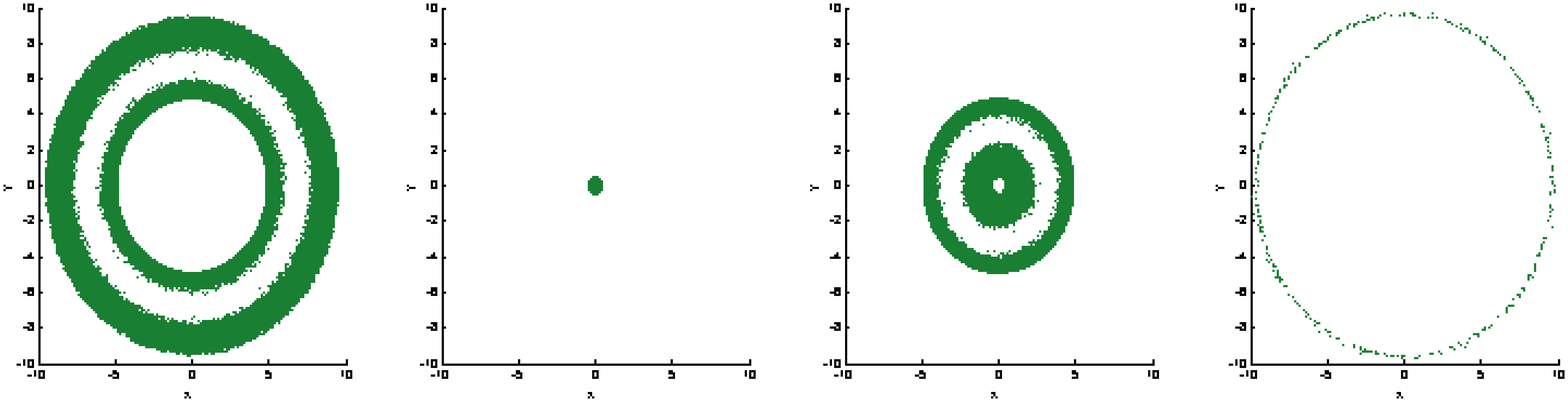}
        } \\
        \subfloat[Iteration 2]{
            \centering
            \includegraphics[height=1.5cm,width=5.1cm]{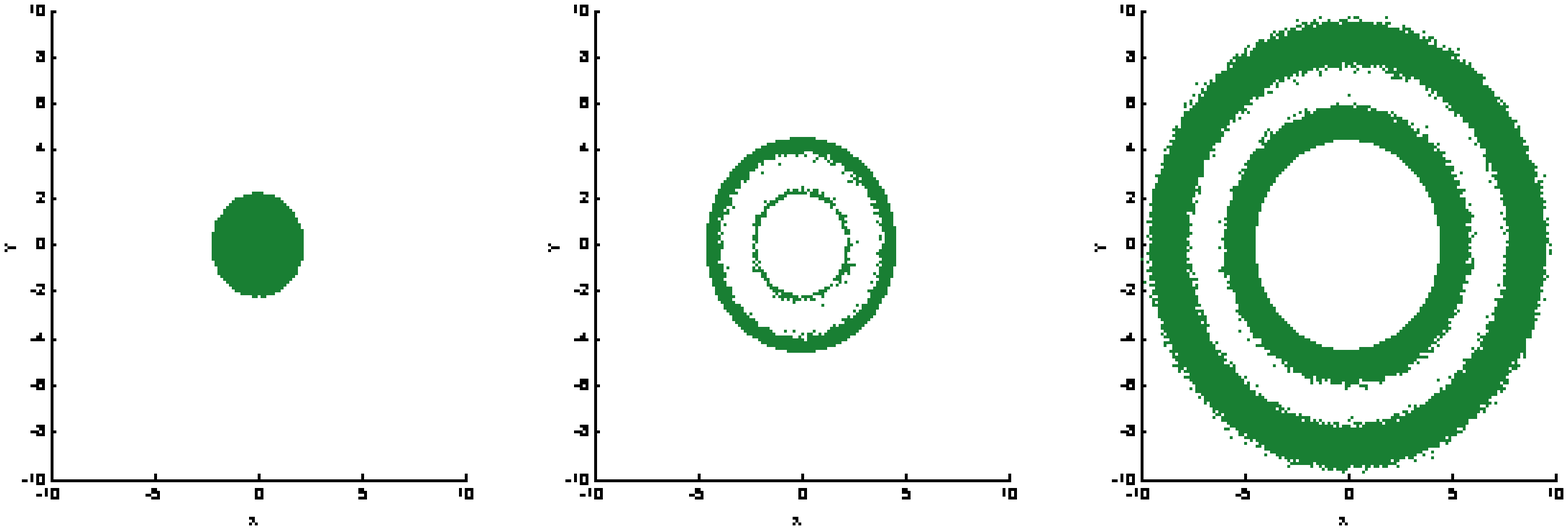}
        } \qquad
        \subfloat[Iteration 3]{
            \centering
            \includegraphics[height=1.5cm,width=5.1cm]{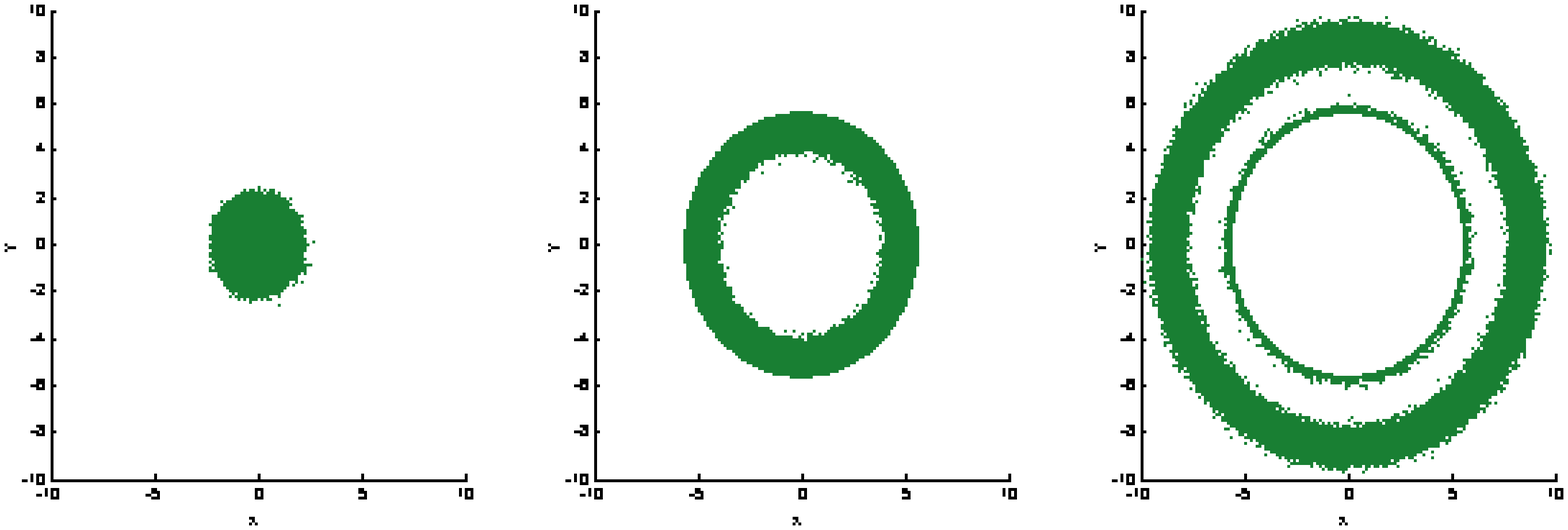}
        } \qquad
        \subfloat[Iteration 4, 5]{
            \centering
            \includegraphics[height=1.5cm,width=5.1cm]{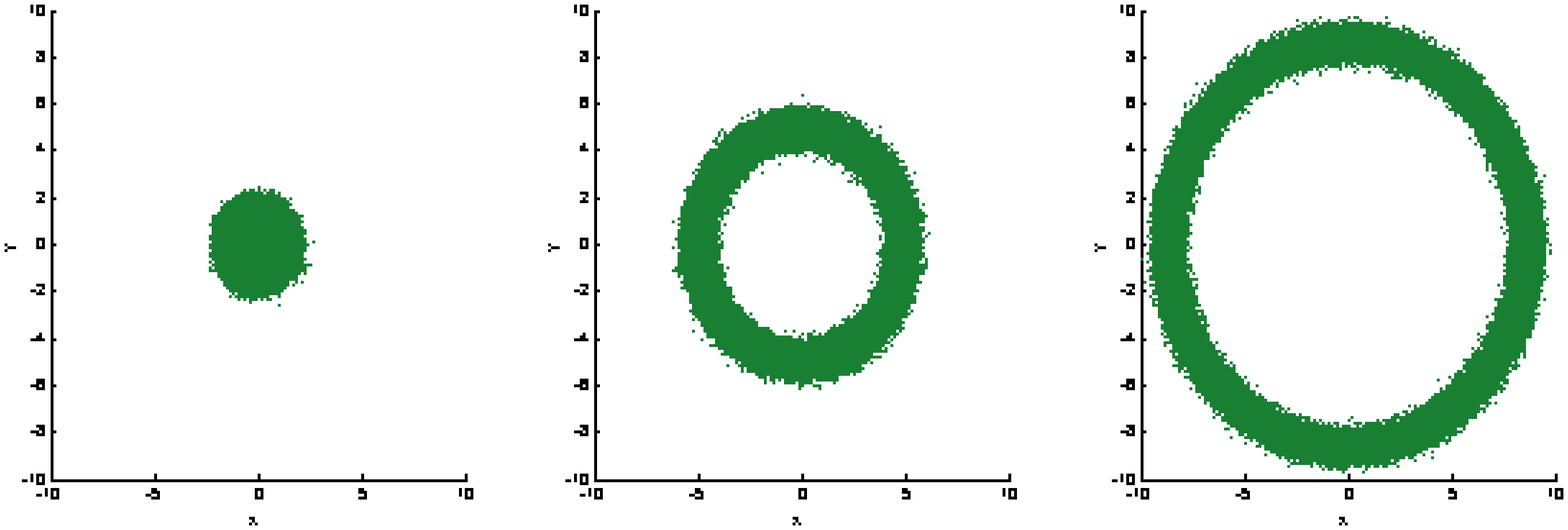}
        }
        \caption{Cluster evolution at each refinement iteration for the concentric ring data with \mi{\lambda_c=24}, \mi{\lambda_g = 3.04\times 10^6}.}
        \label{fig:rings_underseg_refinement_evolution}
    \end{center}
\end{figure*}

\begin{figure}[htpb]
    \begin{center}
        \subfloat[\label{fig:3ring_percPts}]{
            \centering
            \includegraphics[width=0.23\textwidth]{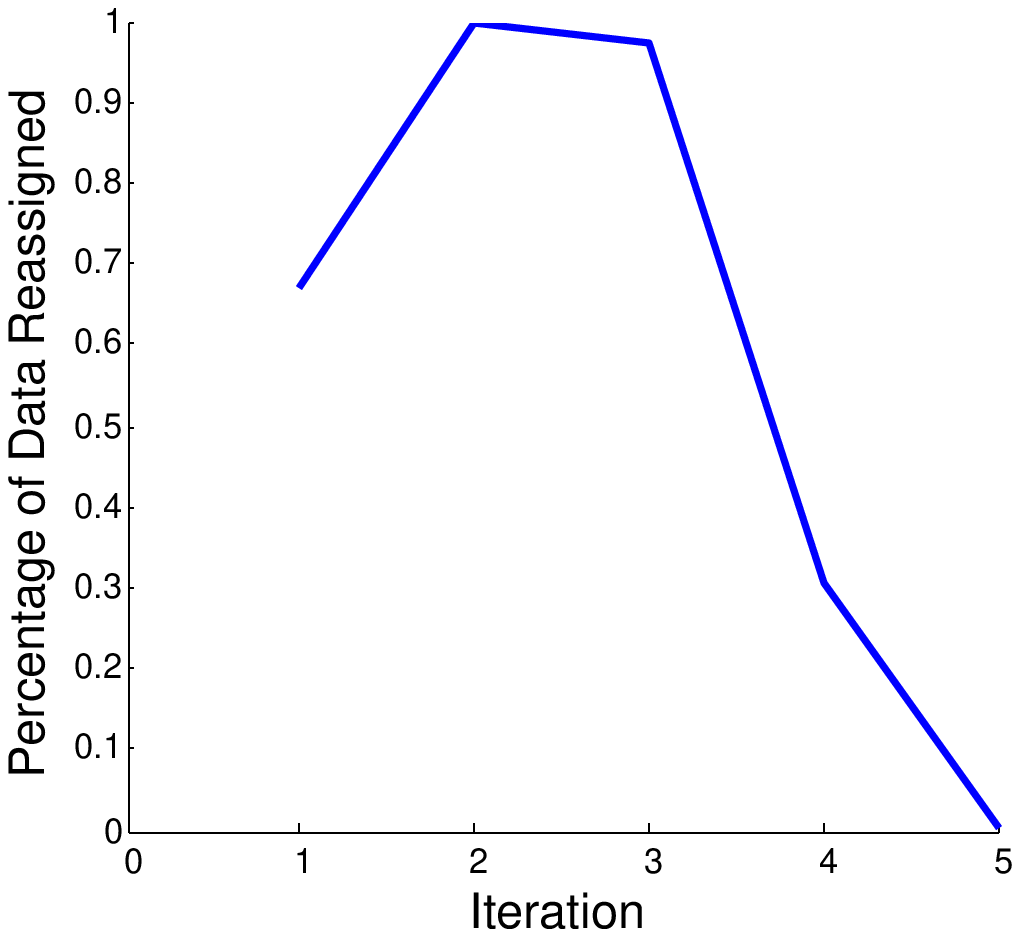}
        }
        \subfloat[\label{fig:3ring_relEnergy}]{
            \centering
            \includegraphics[width=0.23\textwidth]{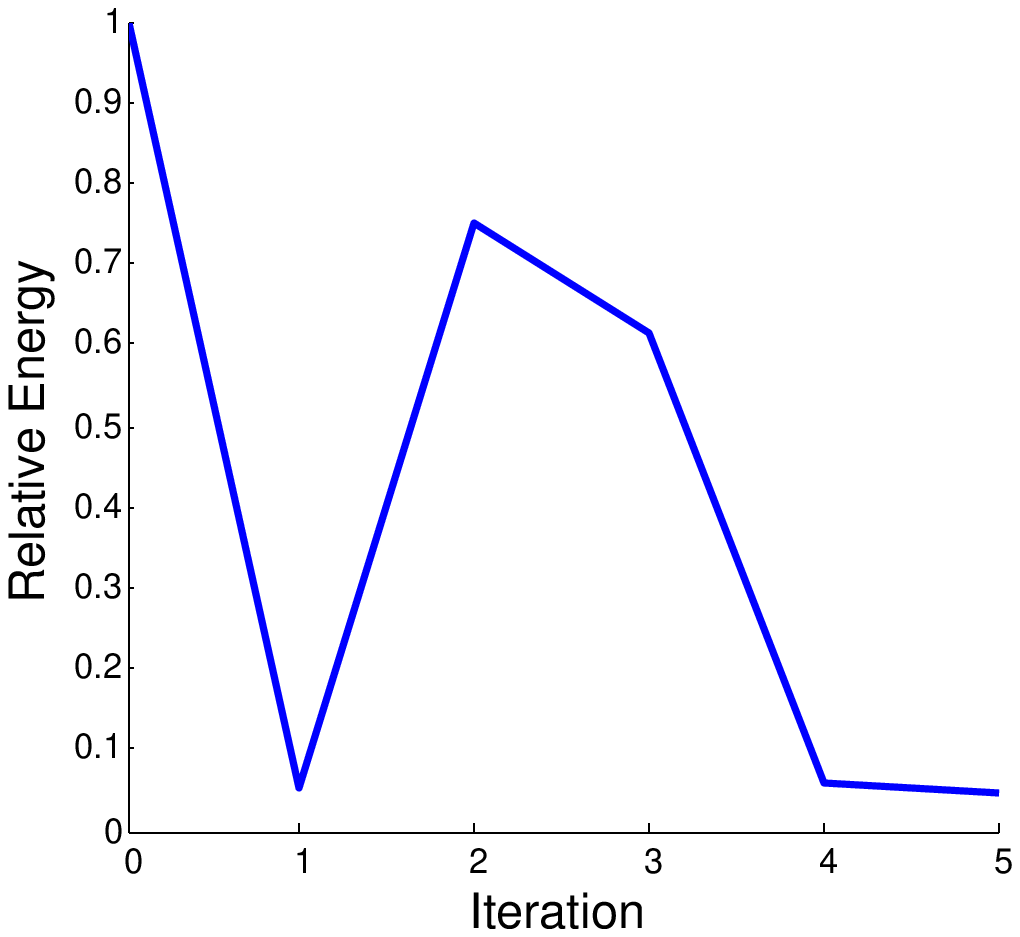}
        }
        \caption{Convergence of the refinement step in clustering the data set from Figure \ref{fig:ring_data} using \mi{\lambda_c=24}.   The percentage of the data set being reassigned (a) and the global energy evaluation relative to the energy prior to refinement (b) are computed for each refinement iteration.}
        \label{fig:3ring_refinement_convergence}
    \end{center}
\end{figure}

To study the convergence properties of the refinement iteration, consider the clustering of the data set in Figure~\ref{fig:2d_poorsep}.  For the parameter choices \mi{\lambda_c = 0.05, 0.1, 0.5}, Figure~\ref{fig:2d_poorsep_refine_converge} shows the computed global energy and the percentage of data being reassigned at each iteration.  For each value of \mi{\lambda_c}, \mi{\lambda_g} is chosen so that the final partition is the four-cluster solution in Figure~\subref*{fig:2d_poorsep_refine}.  Because the value of \mi{\lambda_g} changes between experiments, the absolute energy values in the different experiments cannot be directly compared; however, Figure~\subref*{fig:2d_poorsep_convergence_percpts} demonstrates that larger values of \mi{\lambda_c} tend to result in more points which must be reassigned during the refinement.  The refinement procedure may not converge monotonically, particularly for large values of \mi{\lambda_c}.  When data are merged more aggressively during the initial clustering, the number of misclassified points remaining after the grouping step increases and the refinement step will require more iterations for convergence.  In this experiment, the refinement converges in five iterations for \mi{\lambda_c=0.05}, six iterations for \mi{\lambda_c=0.1}, and ten iterations for \mi{\lambda_c=0.5}.

In clustering the data in Figure~\ref{fig:ring_data} the computational cost of the refinement step is considerably higher for \mi{\lambda_c=24} than for the other cases.  While the computational time of the refinement step increases as the number of clusters increases, it also depends on the stability of the clusters.  With \mi{\lambda_c=24}, the parallel initial clustering does not separate the data well and a great deal of work must be performed during the refinement step in order to obtain the \mi{k=3} solution.
Figures~\ref{fig:rings_underseg_refinement_evolution} and \ref{fig:3ring_refinement_convergence} show the evolution of the cluster configuration for \mi{\lambda_c=24} throughout the refinement step.  Prior to refinement (Figure~\subref*{fig:ring_underseg_groups}) the energy is high (iteration 0 in Figure~\subref*{fig:3ring_relEnergy}), but the first refinement iteration reduces the energy substantially while reassigning nearly \mi{65\%} of the data points and eliminating one cluster.  After the large change in cluster centroid locations, there are many additional points which could be moved to produce a lower energy.  The second iteration eliminates a second cluster, which moves the configuration toward the more-stable \mi{k=3} solution.  From this state, the refinement process proceeds smoothly toward the final configuration, terminating after iteration 5.  The computational cost is very high for refinement iterations in which a large percentage of the data is reassigned.

\subsection{Computational Efficiency} \label{sec:computationalEfficiency}

\begin{figure*}[htpb]
    \begin{center}
        \subfloat[\label{fig:2d_poorsep_time_linear}]{
            \centering
            \includegraphics[width=0.25\textwidth]{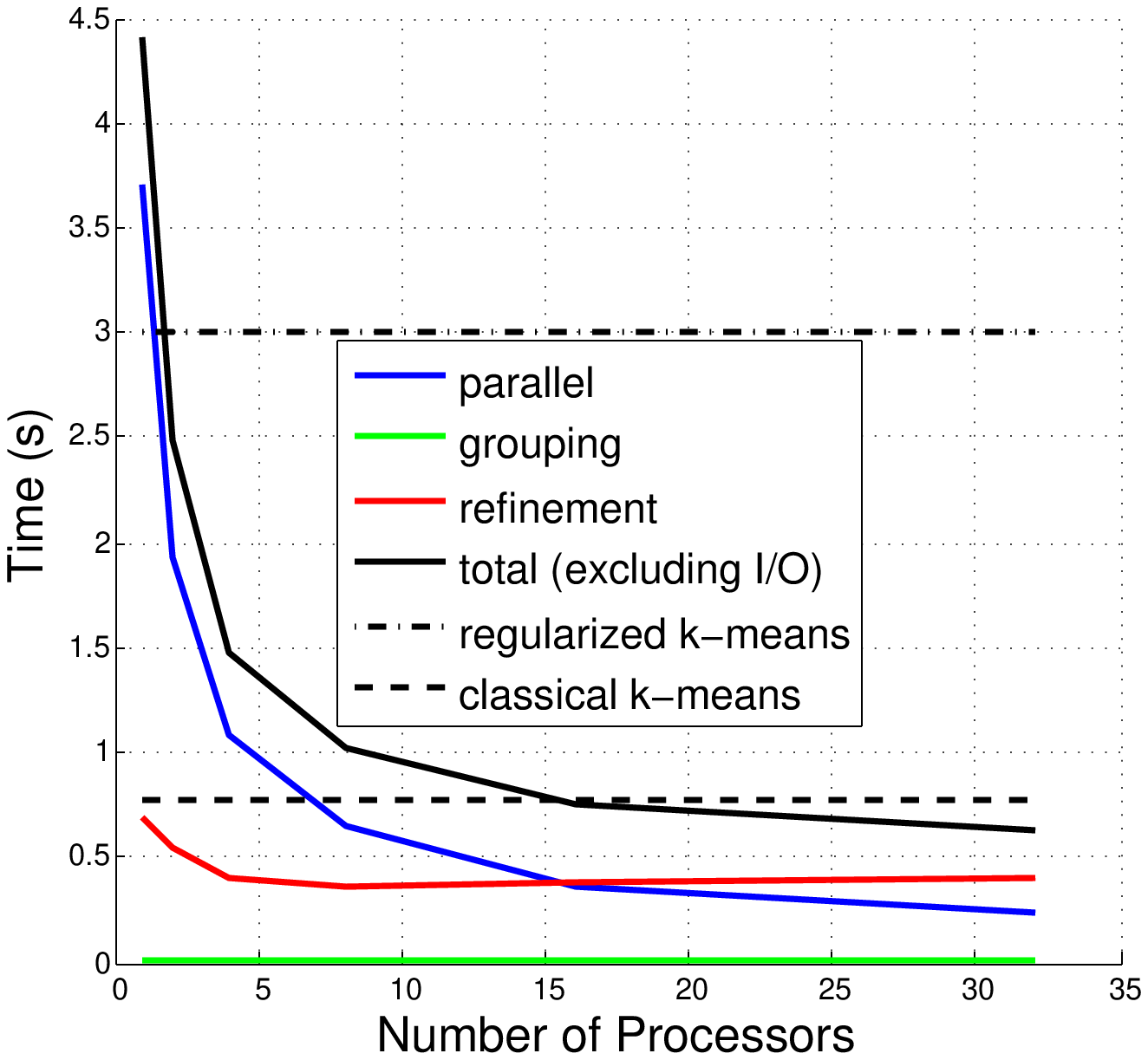}
        }\quad
        \subfloat[\label{fig:2d_poorsep_time_log}]{
            \centering
            \includegraphics[width=0.25\textwidth]{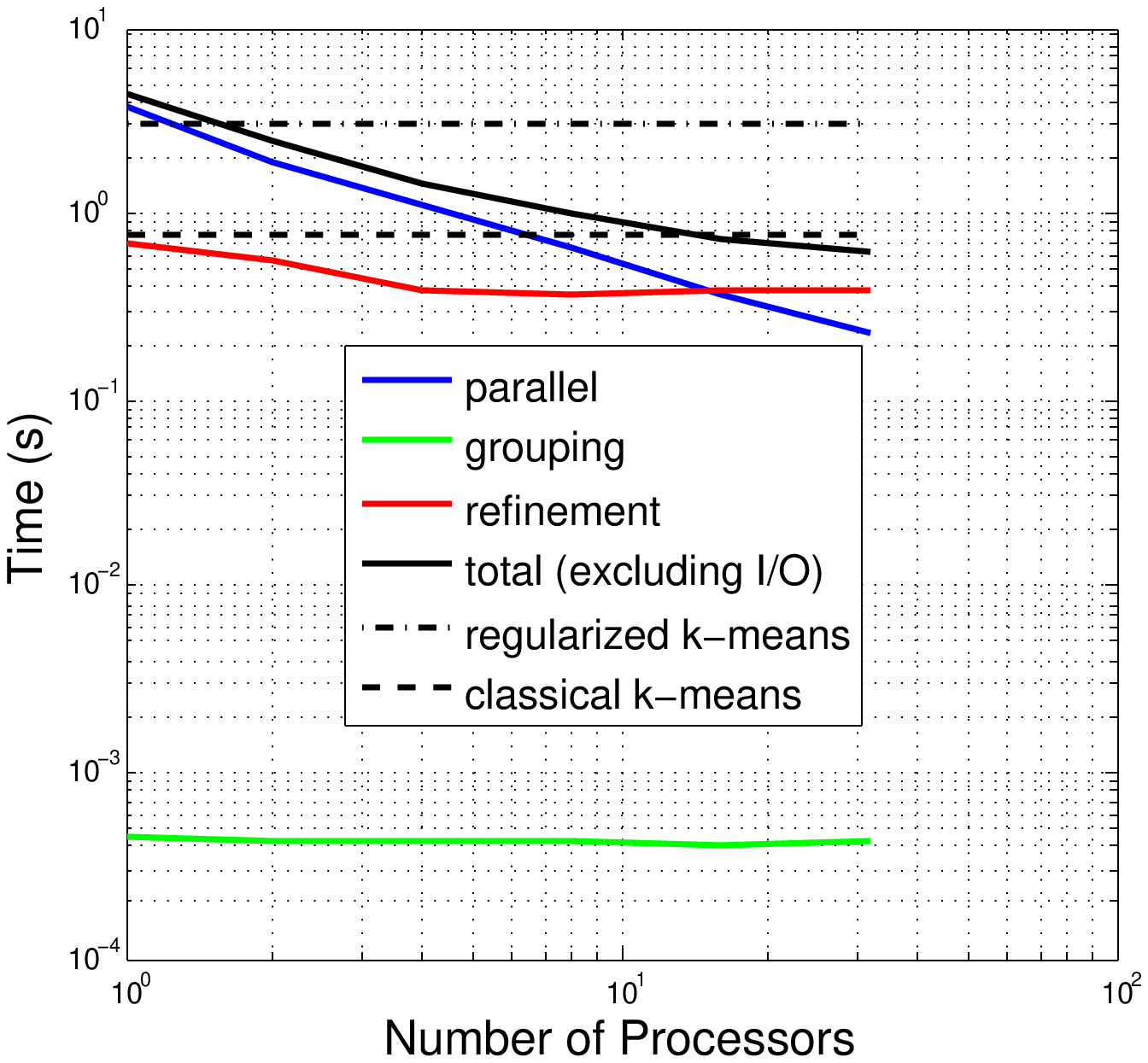}
        }\quad
        \subfloat[\label{fig:2d_poorsep_time_bar}]{
            \centering
            \includegraphics[width=0.25\textwidth]{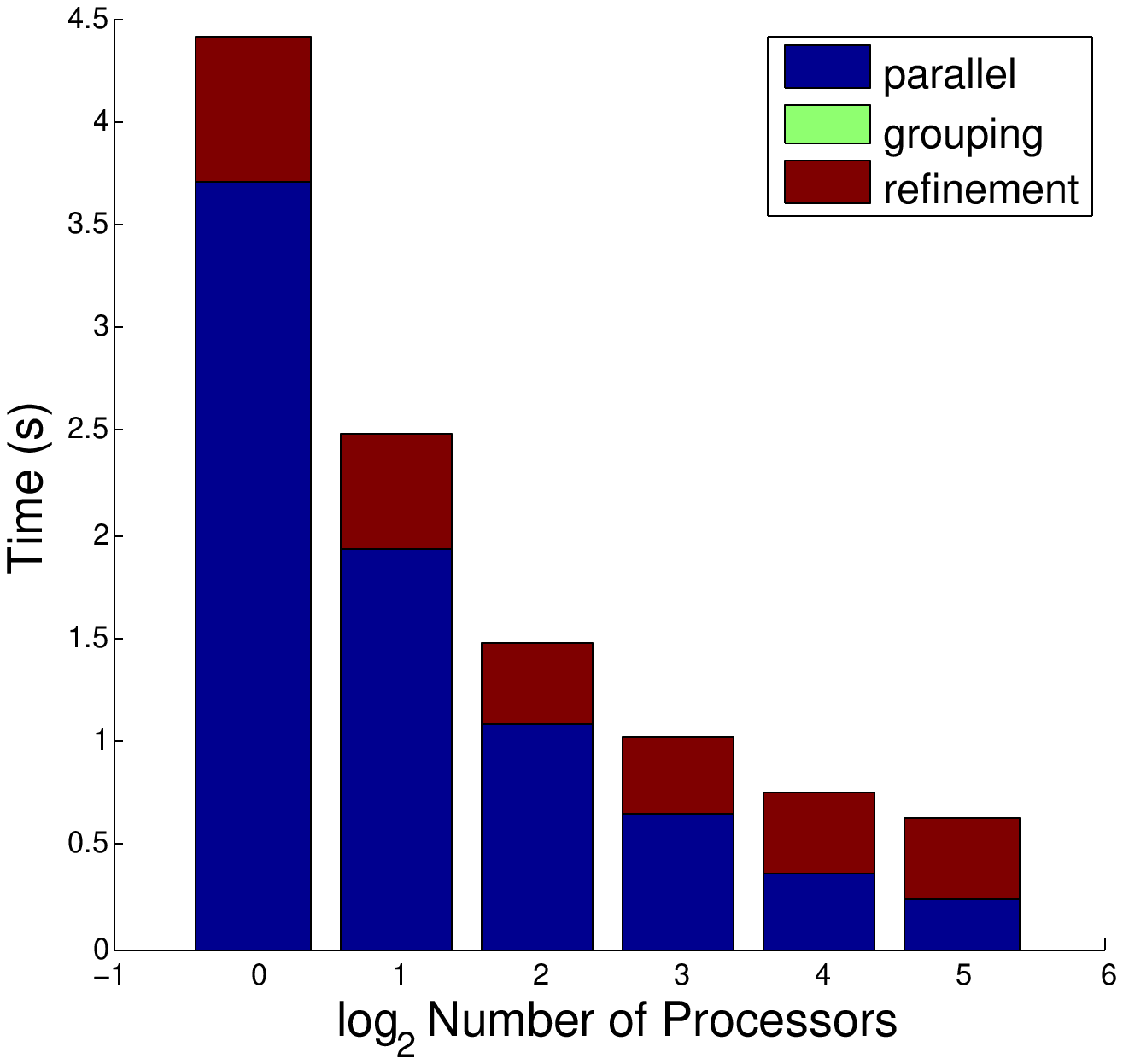}
        }
    \end{center}
    \caption{Scaling of computational time for the parallel adaptive clustering algorithm as the number of processors increases measured on \protect\subref*{fig:2d_poorsep_time_linear} linear and \protect\subref*{fig:2d_poorsep_time_log} logarithmic scale.  For reference, computational time required by classical \mi{k}-means and regularized \mi{k}-means (both running sequentially) are included. The contribution of each individual step to the total computational time is shown in \protect\subref{fig:2d_poorsep_time_bar}.}
    \label{fig:2d_poorsep_efficiency}
\end{figure*}

The scaling performance of the algorithm is analyzed using a shared-memory computing node with 32 processors.  The data set from Figure~\subref*{fig:2d_poorsep_data} is clustered using 128 threads with parameter values \mi{\lambda_c=0.06}, \mi{\epsilon=0.01} (\mi{\lambda_g=418932}).  Figure~\ref{fig:2d_poorsep_efficiency} shows the computational cost of the entire procedure (excluding input and output) as the number of processors is doubled.  All of these experiments produce the \mi{k=4} solution depicted in Figure~\subref*{fig:2d_poorsep_refine}.

The computational cost of the parallel step scales well with the number of processors (Figure~\subref*{fig:2d_poorsep_time_linear},\subref*{fig:2d_poorsep_time_log}).  The computational time for the refinement step scales poorly with the number of processors, quickly becoming constant due to the fact that it exploits only up to \mi{k}-fold parallelism.  While the grouping step does not scale with the number of processors, it occupies an insignificant portion of the total processing time (Figure~\subref*{fig:2d_poorsep_time_bar}). The total execution time for the algorithm (excluding input and output operations) is dominated by the time required for the initial parallel clustering step when the number of processors is small, but dominated by the time required for the refinement step when the number of processors is large.

\begin{figure}[htpb]
    \begin{center}
        \includegraphics[width=0.3\textwidth]{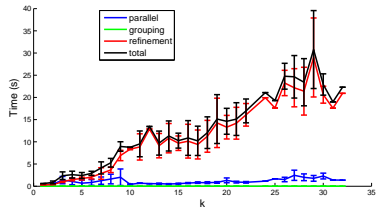}
        \caption{Computational Time (s) vs. the Number of Clusters, \mi{k}}
        \label{fig:2d_poorsep_t_vs_k}
    \end{center}
\end{figure}

In Figure \ref{fig:2d_poorsep_t_vs_k}, the number of processors is fixed at 16 and the data from Figure \ref{fig:2d_poorsep} is clustered many times with different parameter choices.  In particular, we select \mi{\lambda_c} from a set of 6 uniformly spaced points in the range \mi{\left\lbrack 0.05,0.1 \right\rbrack} and choose \mi{\epsilon} from a set of 100 uniformly spaced points in the range \mi{\left\lbrack 0.003,0.015 \right\rbrack} for a total of 600 parameter combinations.  For each set of parameters clustering is performed 25 times, randomly reordering the data each time.  These experiments resulted in partitions with \mi{1 \leq k \leq 32}, though \mi{k<5} occurred much more frequently than configurations with higher numbers of clusters.
Figure~\ref{fig:2d_poorsep_t_vs_k} demonstrates that the refinement step is the primary driver of the overall computational cost, and \mi{k} drives the cost of the refinement step.  For each value of \mi{k} the height of the curve is the time required for the algorithm, averaged over each trial that resulted in \mi{k} clusters, with vertical whiskers illustrating the standard deviation from the average.  The computational cost of the parallel step is generally low across the experiments, with small variance across repeated trials.  The computational cost of the grouping step has an insignificant computational cost with negligible variance.  However, the refinement step has a computational cost which increases significantly as \mi{k} increases.  The variance is larger for larger values of \mi{k} due to the small number of experiments resulting in \mi{k>5}.

\subsection{Stability}

\begin{figure}[htpb]
    \begin{center}
        \subfloat[\mi{\lambda_c=0.06}\label{fig:2d_poorsep_k_vs_lambda006}]{
            \includegraphics[width=0.23\textwidth]{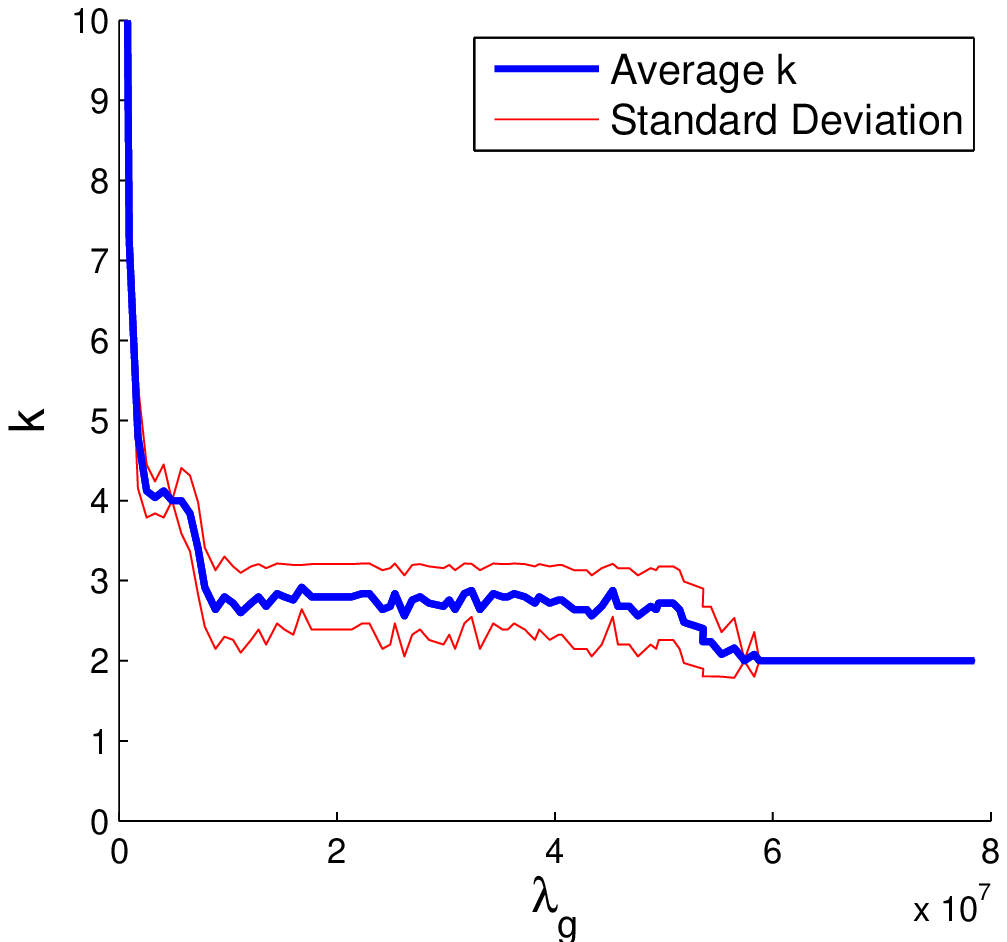}
        }
        \subfloat[\mi{\lambda_c=0.1}\label{fig:2d_poorsep_k_vs_lambda01}]{
            \includegraphics[width=0.23\textwidth]{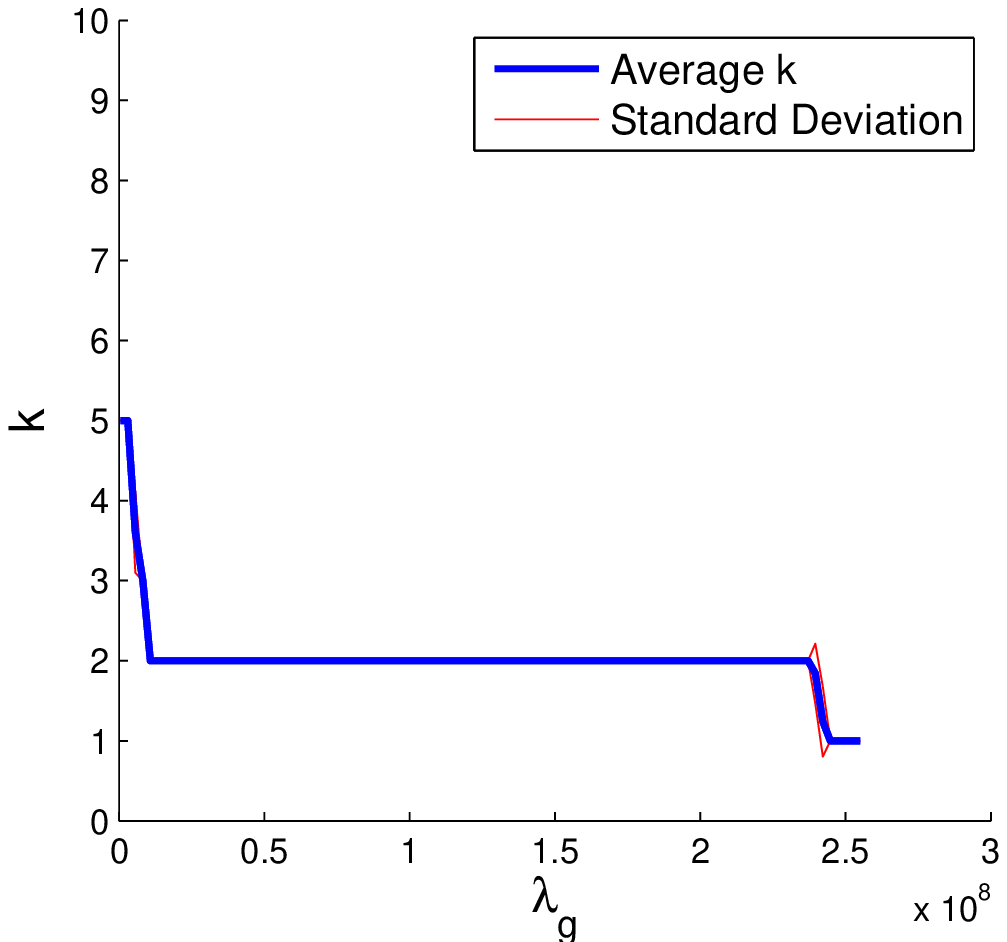}
        }
        \caption{Number of Clusters vs. \mi{\lambda_g}, average over 25 trials with random data order}
        \label{fig:2d_poorsep_k_vs_lambda}
    \end{center}
\end{figure}

The initial ordering of the data and the choice of regularization parameters affect the final cluster configuration.  We repeat the clustering of the data from Figure \ref{fig:2d_poorsep} while varying the data order and parameter values.  Figure~\ref{fig:2d_poorsep_k_vs_lambda} shows the value of \mi{k} for the final cluster sets as a function of \mi{\lambda_g}, corresponding to \mi{\lambda_c=0.06} and \mi{\lambda_c=0.1}.  The red lines in the charts indicate the standard deviation of the observed number of clusters.  In the context of the PAC algorithm, the stability of a cluster configuration depends on its robustness to perturbations both in the parameter values and in the initial order of the data.

The stability of a cluster configuration is assessed by the sensitivity of the solution to the value of the regularization parameter and the initial ordering of the data.  A stable solution is one in which the outcome (\textit{i.e.}, the number of clusters, \mi{k}) changes very little when the input order of the data or the regularization parameter value is changed. Figure~\ref{fig:2d_poorsep_k_vs_lambda} illustrates that the two-cluster solution is more stable than the four-cluster solution.  This is caused by the large separation between the clusters in the two-cluster solution.  Figure~\subref*{fig:2d_poorsep_k_vs_lambda01} shows that the regularization parameter may vary over a large interval without changing the cluster configuration.

In Figure~\subref*{fig:2d_poorsep_k_vs_lambda006}, the regularization parameter is small enough that the large decentralized cluster begins to be subdivided.  However, over a number of trials, the number of clusters produced varies as the input order of the data is perturbed.  The red bars on the plots in Figure~\subref*{fig:2d_poorsep_k_vs_lambda006} and \subref{fig:2d_poorsep_k_vs_lambda01} show the variance of the computed value of \mi{k} during 25 trials with random reordering of the data.  The two-cluster solution is quite stable with respect to the data order, while solutions with more than two clusters are more sensitive to the input order of the data.

\section{Application to Time-Dependent Data} \label{sec:streaming}

We apply the PAC method to a problem where the data set is changing over time.  This is important because many current applications rely on data that arrives continuously, and require that the classification model be capable of adapting to new data as it becomes available. In such cases, the optimal number of clusters to partition the data set may change over time; methods which fix the number of clusters to be constant have a serious drawback in this setting.  A second challenge is that dynamically updating the partition to integrate new data is cumbersome or impossible.  The parallel adaptive clustering algorithm provides a framework that can be naturally extended to clustering data streams, with the capability to dynamically update the cluster configuration to integrate new data and adjust the number of clusters.  This is accomplished by utilizing the intermediate cluster configuration when incorporating new data, as described in Algorithm~\ref{alg:prkmtv}.

\subsection{Time-Dependent PAC Algorithm}

To efficiently cluster time-dependent data, the PAC algorithm keeps a record of the output at each time step.  At time \mi{t}, a new set of data, \mi{\hat{\set{X}}^t}, becomes available and the parallel step is performed only on the new data to form a set of clusters, \mi{\hat{\sset{C}}^t}.  We allow \mi{\sset{C}^t} to denote the collection of the results of the parallel clustering for all time steps up to \mi{t}, so that \mi{\sset{C}^t = \cup_{\tau = 1}^{t} \hat{\sset{C}}^\tau}.  Similarly, we let \mi{\set{X}^t} represent the aggregation of all data that has become available by time step \mi{t}: \mi{\set{X}^t = \cup_{\tau = 1}^{t} \hat{\set{X}}^\tau}.
The global cluster configuration at time \mi{t} is obtained by performing the grouping and refinement procedures on \mi{\sset{C}^t}.


\begin{algorithm}[t]
    \SetInd{0.25em}{0.75em}
	\caption{Parallel Adaptive Clustering for Streaming Data}   	\label{alg:prkmtv}
	\KwIn{\mi{n}; \mi{\lambda_c}; \mi{\lambda_g}; \textit{ITER\_MAX}; \textit{TOL}}
	\KwOut{cluster configuration}
	\KwInit{\mi{t=0}, \mi{\set{X}^0=\setbrace{\emptyset}}, \mi{\sset{C}^0=\setbrace{\emptyset}}}
	\While{more data}{
		Retrieve new data, \mi{\tilde{\set{X}} = \bigcup_{p=1}^n \tilde{\set{X}}_p}; \enskip set \mi{t \leftarrow t + 1}\;
		\ForEach{\mi{p \in \setbrace{1 \dots n}}, in parallel}{
			Use Algorithm \ref{alg:rkm} to partition each subset of the new data, \mi{\tilde{\set{X}}_p}, into a set of clusters, \mi{\tilde{\sset{C}}_p}, by minimizing (\ref{eq:thread_energy})\;
		}
		Collect clusters from each parallel thread, \mi{\tilde{\sset{C}} = \bigcup_{p=1}^{n}\tilde{\sset{C}}_p}\;
		Update \mi{\set{X}^t = \set{X}^{t-1} \bigcup \tilde{\set{X}}}, \hspace{5pt} \mi{\sset{C}^t = \sset{C}^{t-1} \bigcup \tilde{\sset{C}}}\;
		Use Algorithm \ref{alg:rkm} to cluster \mi{\sset{C}^t} into groups, \mi{\sset{G}}, by minimizing (\ref{eq:group_energy}), \;
		Compute refined clusters for time \mi{t} using Algorithm \ref{alg:refinement};
	}
\end{algorithm}

When the number of data arriving at each incremental time step is roughly constant, it is appropriate to use a constant \mi{\lambda_c} for all time steps. \mi{\lambda_g} can be adjusted at each time step to account for the increasing size of the aggregate data set.  As \mi{\size{\set{X}}} grows \mi{\size{\sset{C}}} increases even though the average size of \mi{\set{C}\in\sset{C}} may not change, causing the regularization term to lose significance compared to the fitting term when \mi{\lambda_g} is constant.  In the time-varying case, \mi{\lambda_g} is updated to account for both the sizes of the clusters (Theorem~\ref{thm:lambda_and_cluster_size}) and changes in \mi{\size{\set{X}}}.
\begin{equation}  \label{eq:time_varying_lambda}
     \lambda_g^t = \epsilon\left(\dfrac{\size{\set{X}^t}}{\size{\sset{C}^t}}\right)^2 \left(\dfrac{\size{\set{X}^t}}{\size{\set{X}^1}}\right)^{\nu} \,.
\end{equation}
The cluster configuration produced by the time-dependent PAC at time \mi{t} with \mi{n} parallel threads is equivalent to the result of the PAC algorithm on the aggregate data \mi{\set{X}^t} with \mi{nt} parallel threads.  In the following experiments \mi{\nu=0.1} is used, but practice we have found that \mi{0 < \nu < 1} may be appropriate.

\subsection{Experiment for Data Stream Clustering}

\begin{figure*}[htpb]
    \begin{center}
         \subfloat[2-D time-varying data set\label{fig:2d_time_varying_data_full}]{
             \includegraphics[width=0.25\textwidth]{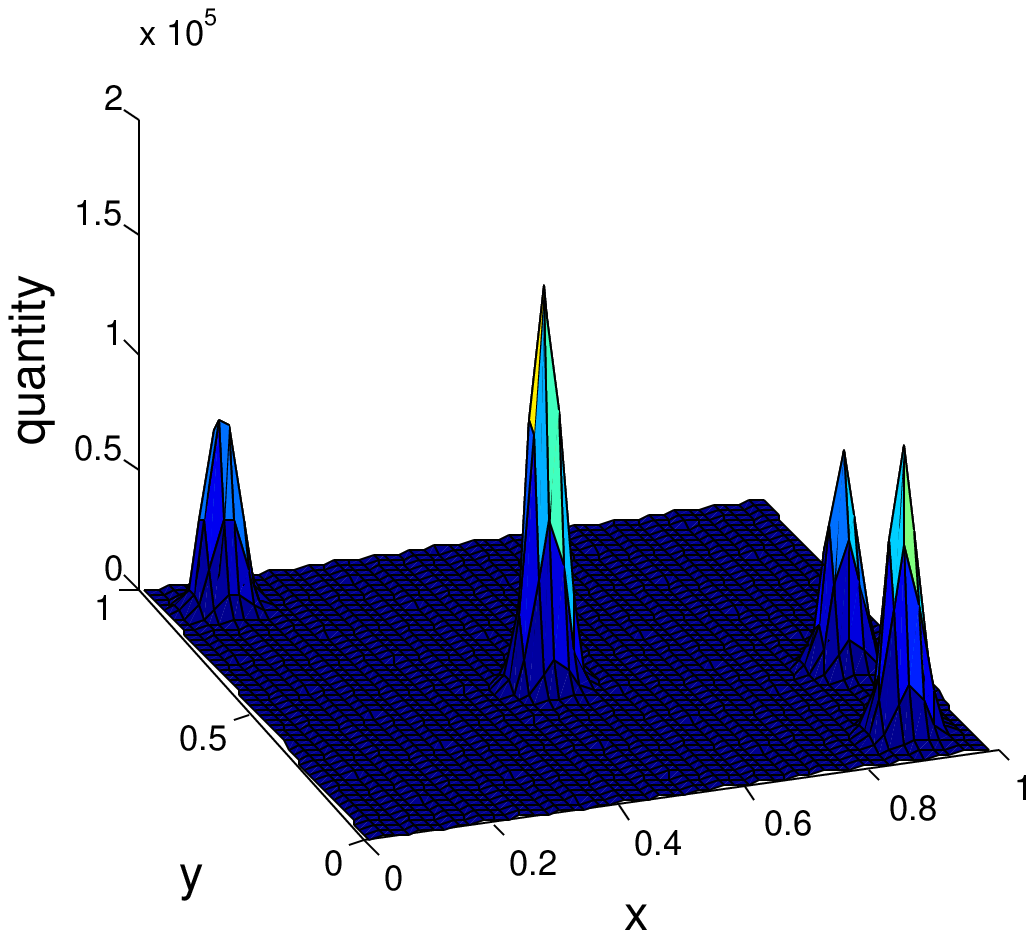}
        } \quad
        \subfloat[Accumulation of incrementally arriving data\label{fig:2d_time_varying_data_increment}]{
            \includegraphics[width=0.33\textwidth]{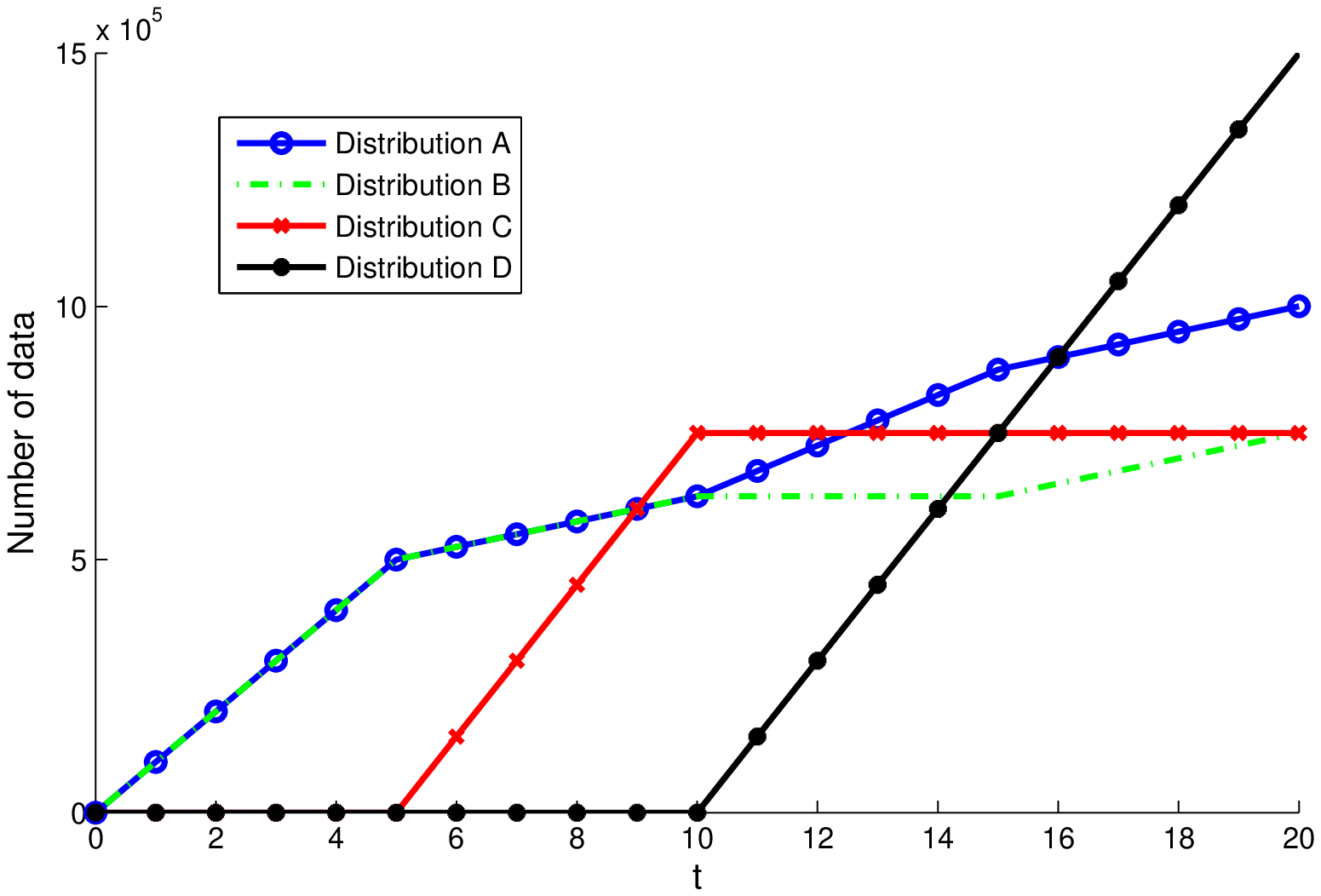}
        } \quad
        \subfloat[Number of computed clusters for each time step\label{fig:2d_time_varying_clusters}]{
            \includegraphics[width=0.33\textwidth]{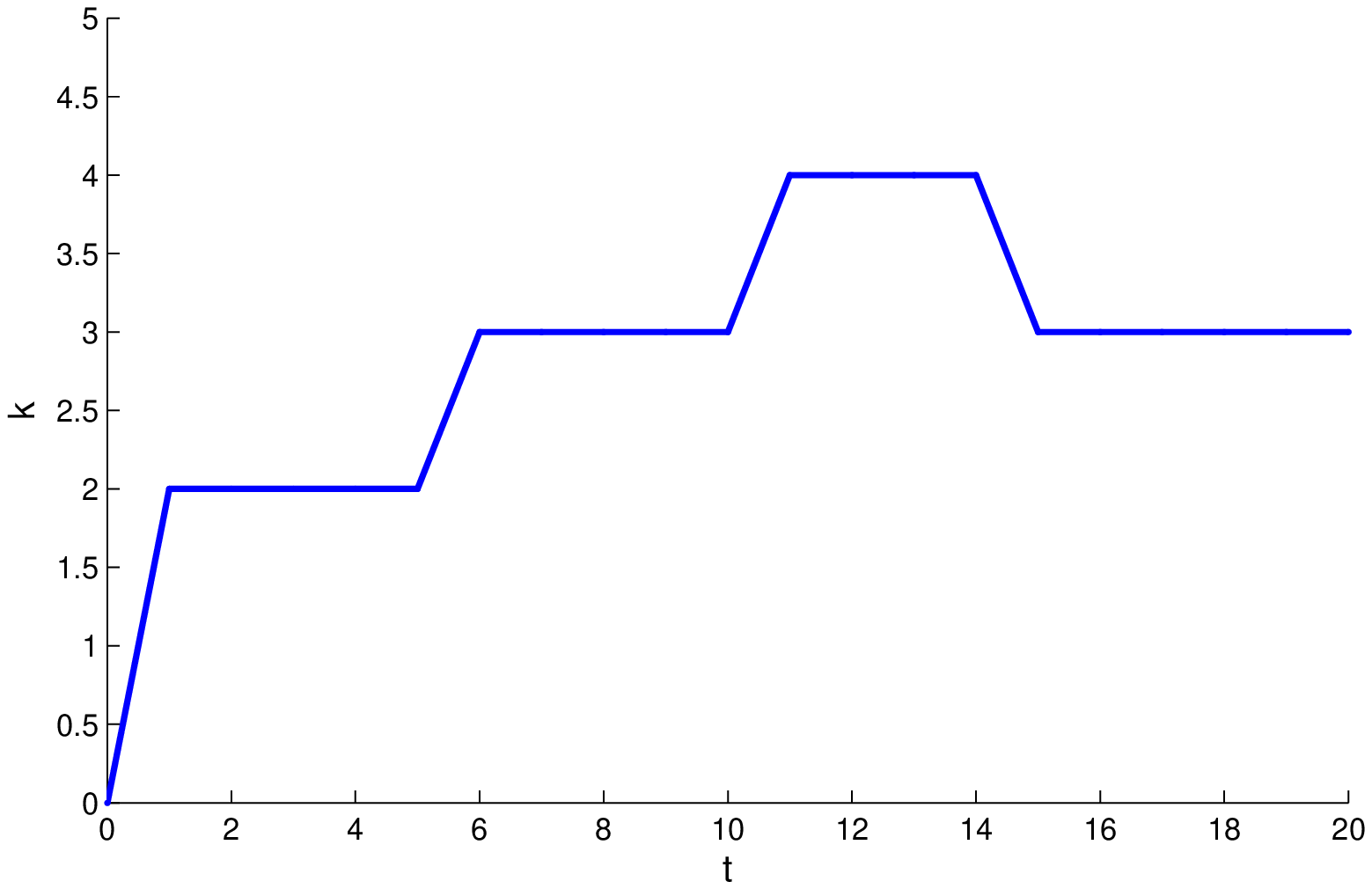}
        }
        \caption{Time-varying accumulation of data set and contribution from underlying data distributions, with corresponding number of clusters at each time step.}
        \label{fig:incrementalData}
    \end{center}
\end{figure*}

To illustrate the clustering of time-dependent data using the PAC algorithm, we consider a data set containing points drawn at random from four distinct Gaussian distributions with different means but similar variance (Figure~\subref*{fig:2d_time_varying_data_full}).  We label each of these four Gaussian peaks as follows: subset A is centered near \mi{(0.9,0.1)}; subset B is centered near \mi{(0.1,0.9)}; subset C is centered near \mi{(0.9,0.4)}; and subset D is centered near \mi{(0.5,0.5)}.  This data set arrives incrementally over a period of 20 time steps, but the distribution of data arriving at each time step changes over time.  

Figure~\subref*{fig:2d_time_varying_data_increment} illustrates the amount of data from each subset which has arrived by time \mi{t}.  During the first few iterations, only data from subsets A and B arrive.  Then, from \mi{t=6} to \mi{t=10}, most of the new data is from subset C, with only a small amount of new data from A and B.  From \mi{t=11} until the final time, nearly all of the new data is from subset D.

\begin{figure*}[htpb]
  \begin{center}
      \subfloat[New Data Arriving, \mi{t=4}]{
        \centering
        \includegraphics[width=0.22\textwidth]{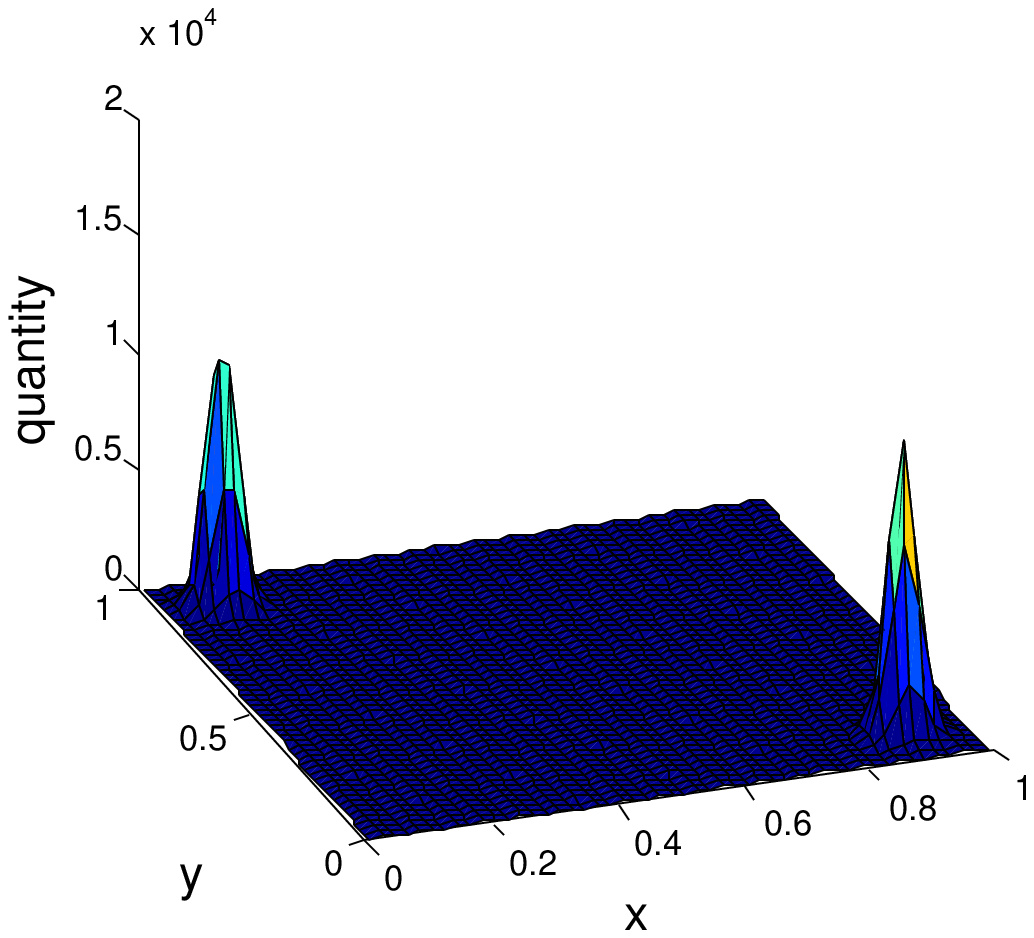}
      }
      \subfloat[New Data Arriving, \mi{t=8}]{
        \centering
        \includegraphics[width=0.22\textwidth]{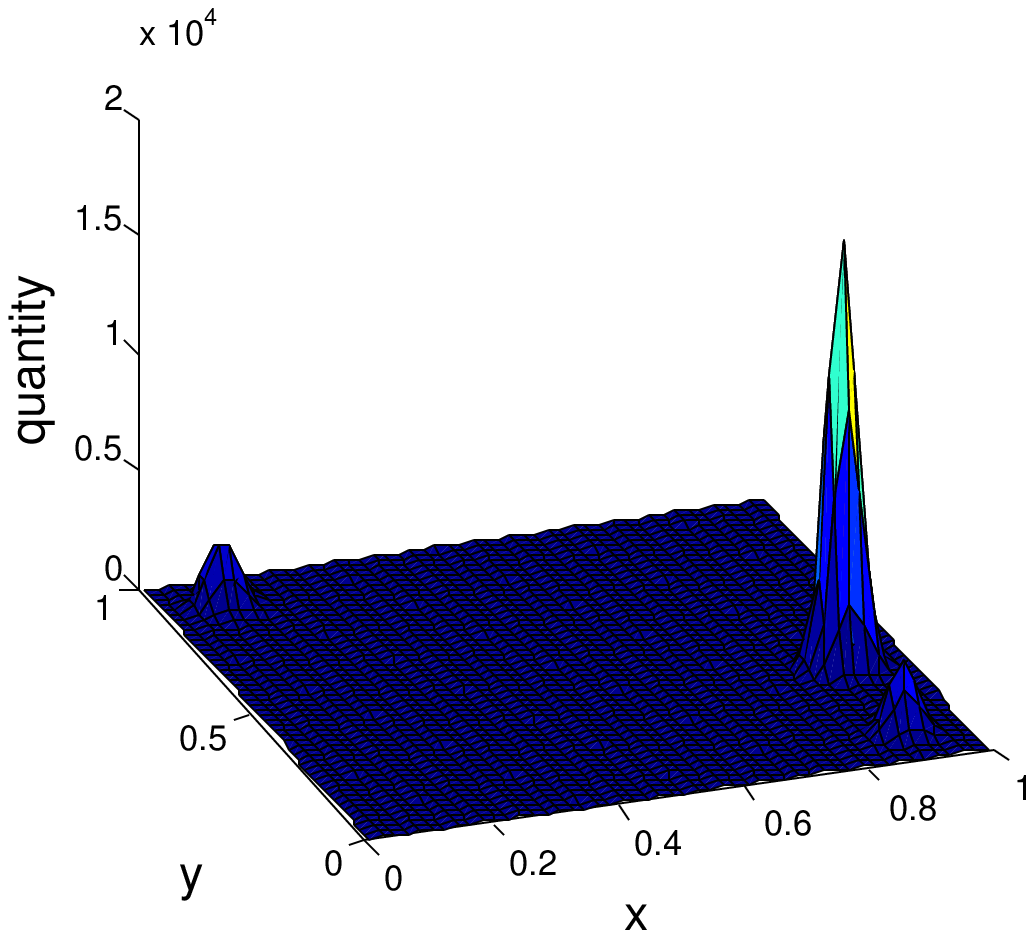}
      }
      \subfloat[New Data Arriving, \mi{t=12}]{
        \centering
        \includegraphics[width=0.22\textwidth]{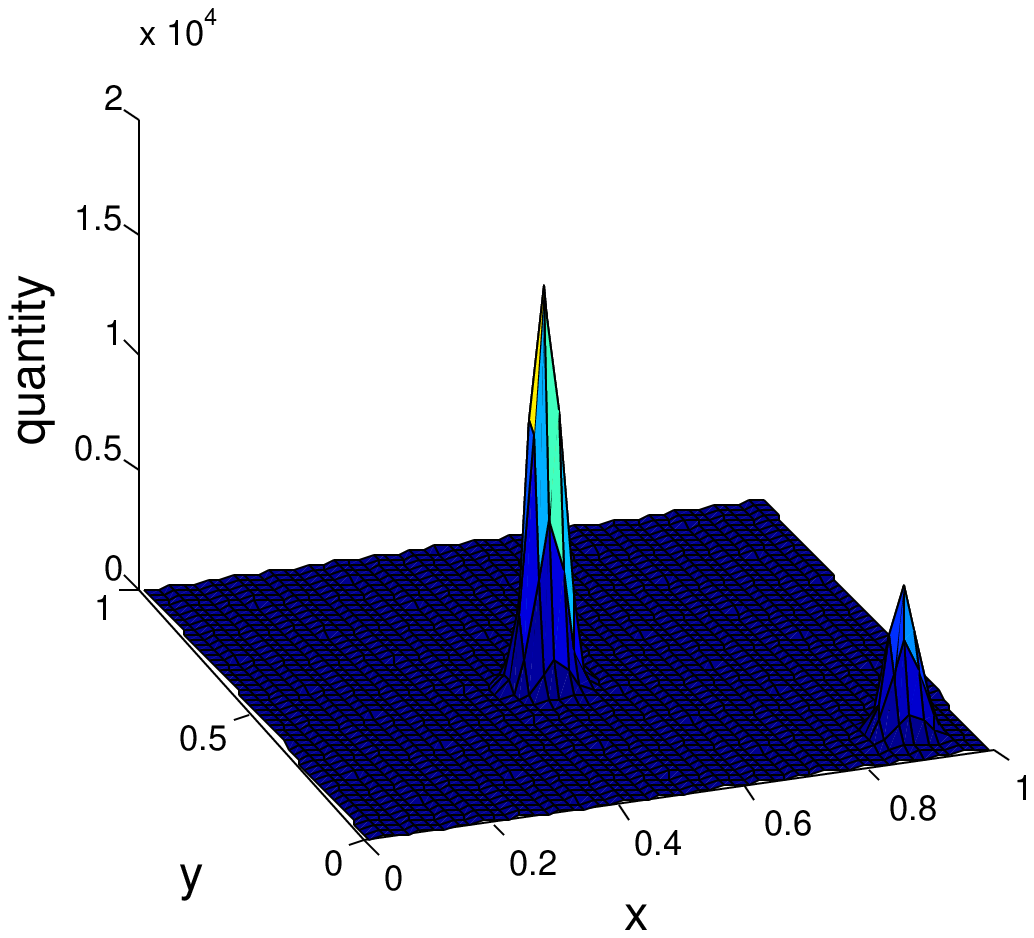}
      }
      \subfloat[New Data Arriving, \mi{t=16}]{
        \centering
        \includegraphics[width=0.22\textwidth]{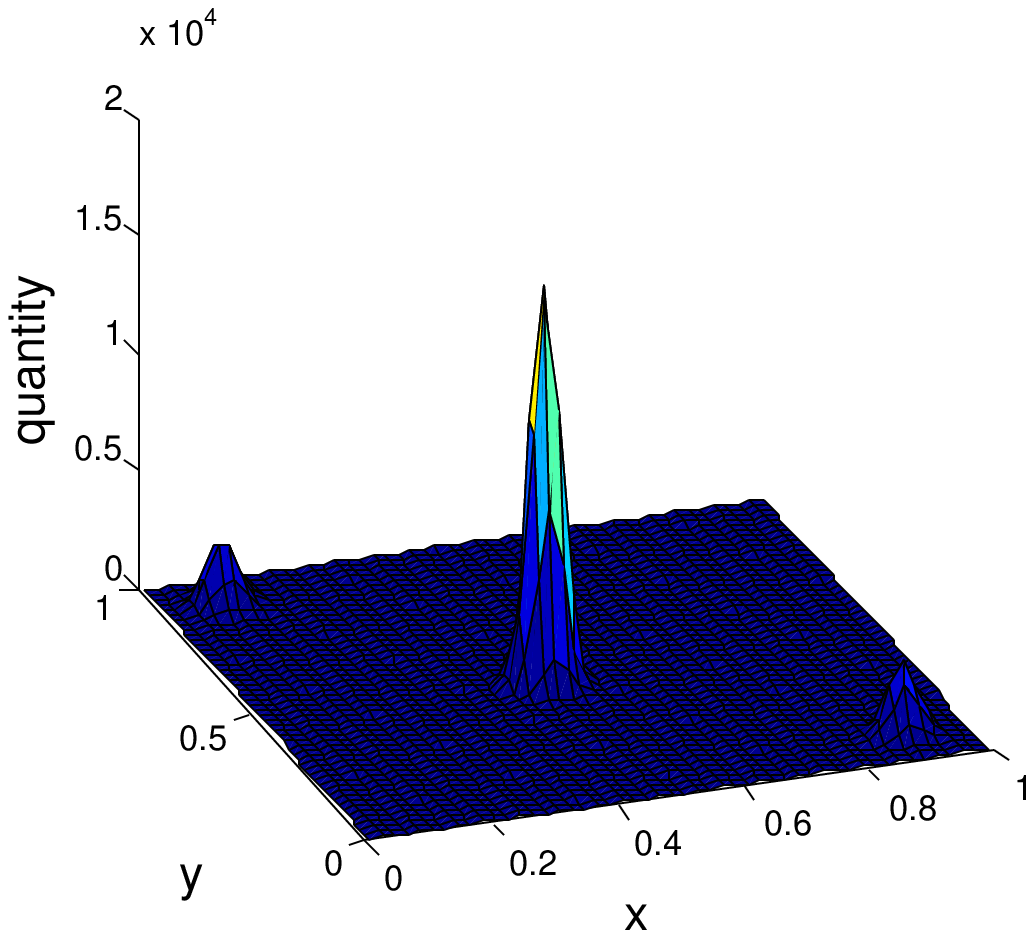}
      } \\

      \subfloat[Aggregate Data, \mi{t=4}]{
        \centering
        \includegraphics[width=0.22\textwidth]{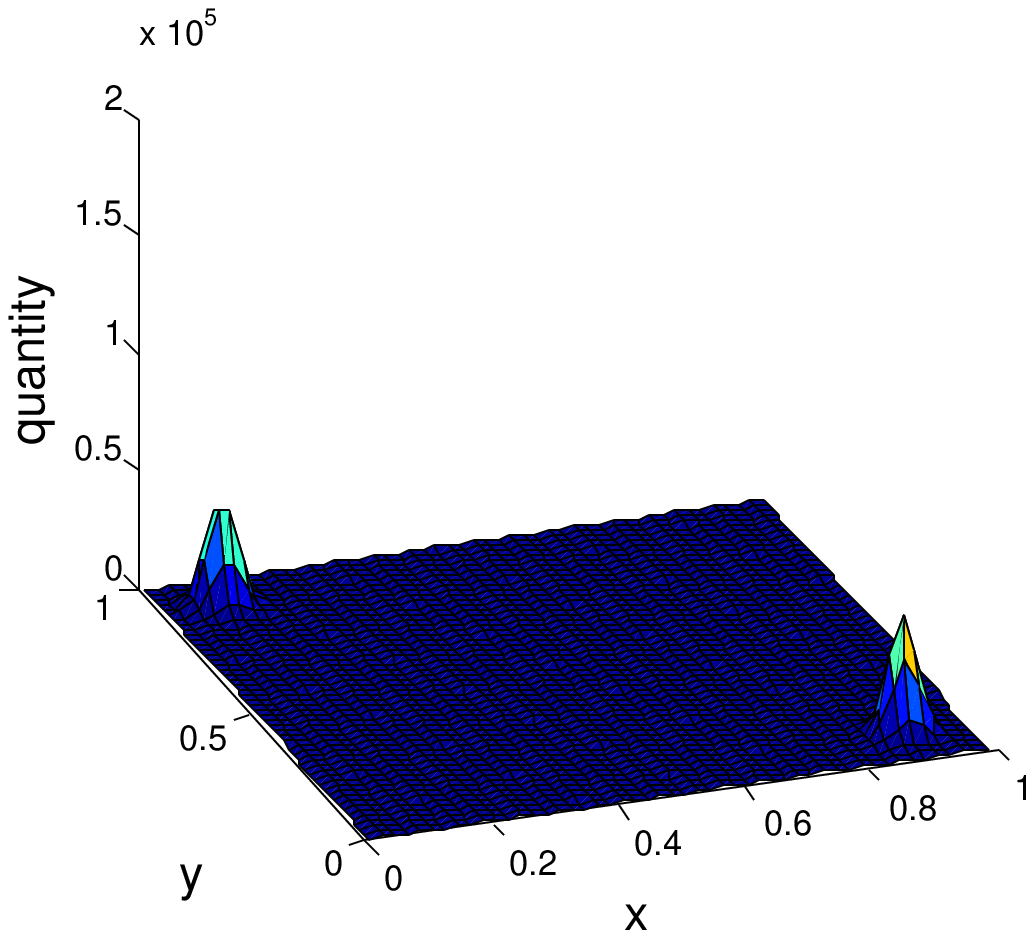}
      }
      \subfloat[Aggregate Data, \mi{t=8}]{
        \centering
        \includegraphics[width=0.22\textwidth]{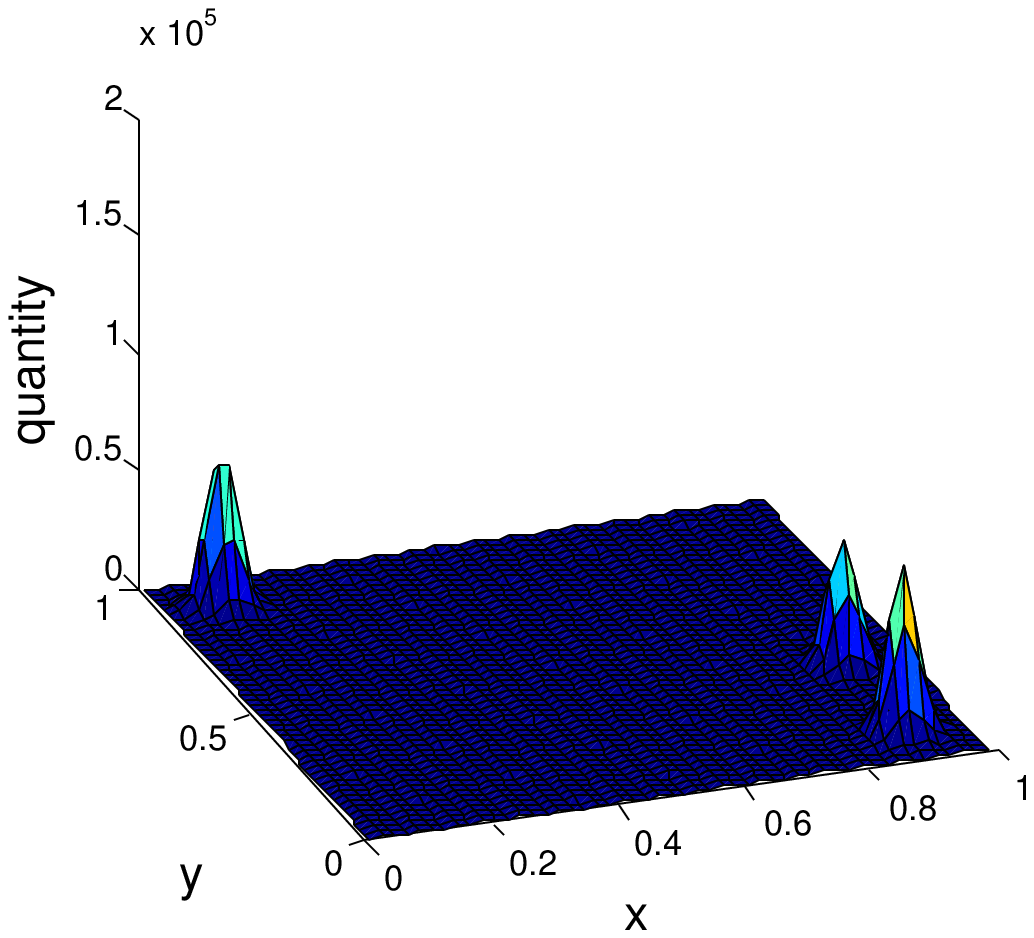}
      }
      \subfloat[Aggregate Data, \mi{t=12}]{
        \centering
        \includegraphics[width=0.22\textwidth]{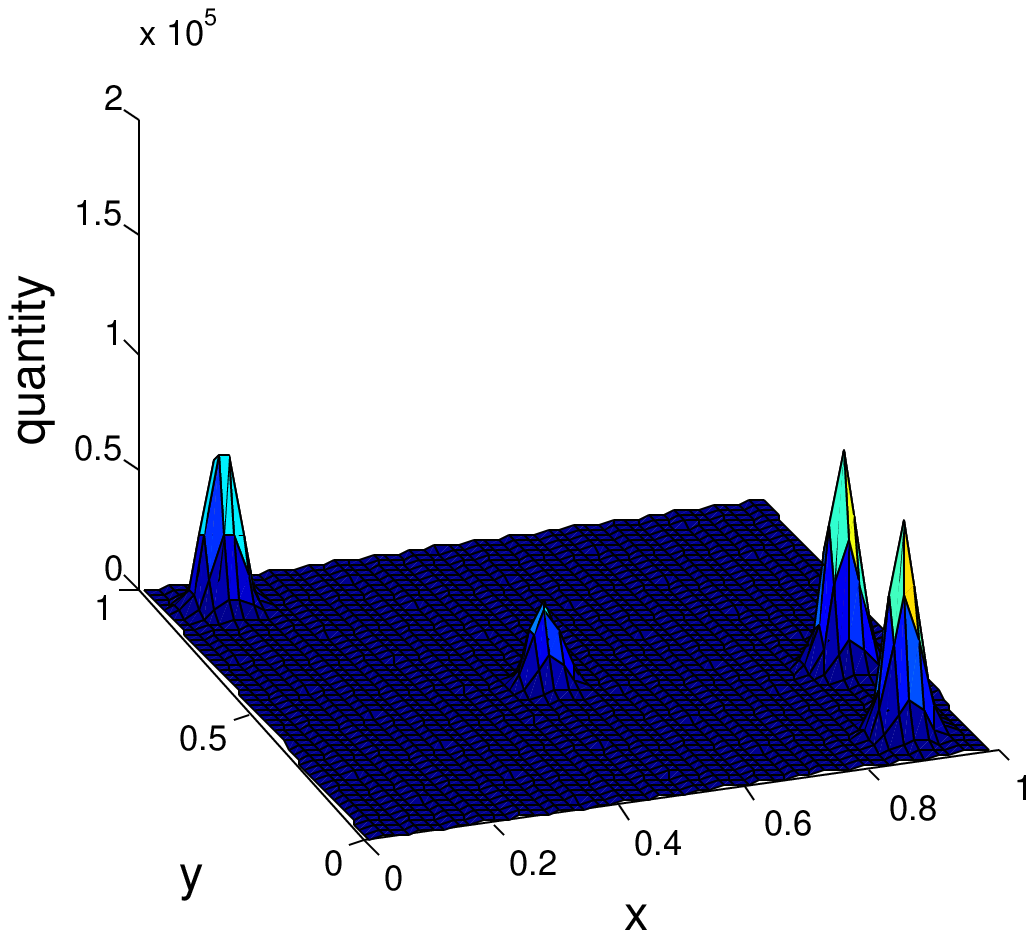}
      }
      \subfloat[Aggregate Data, \mi{t=16}]{
        \centering
        \includegraphics[width=0.22\textwidth]{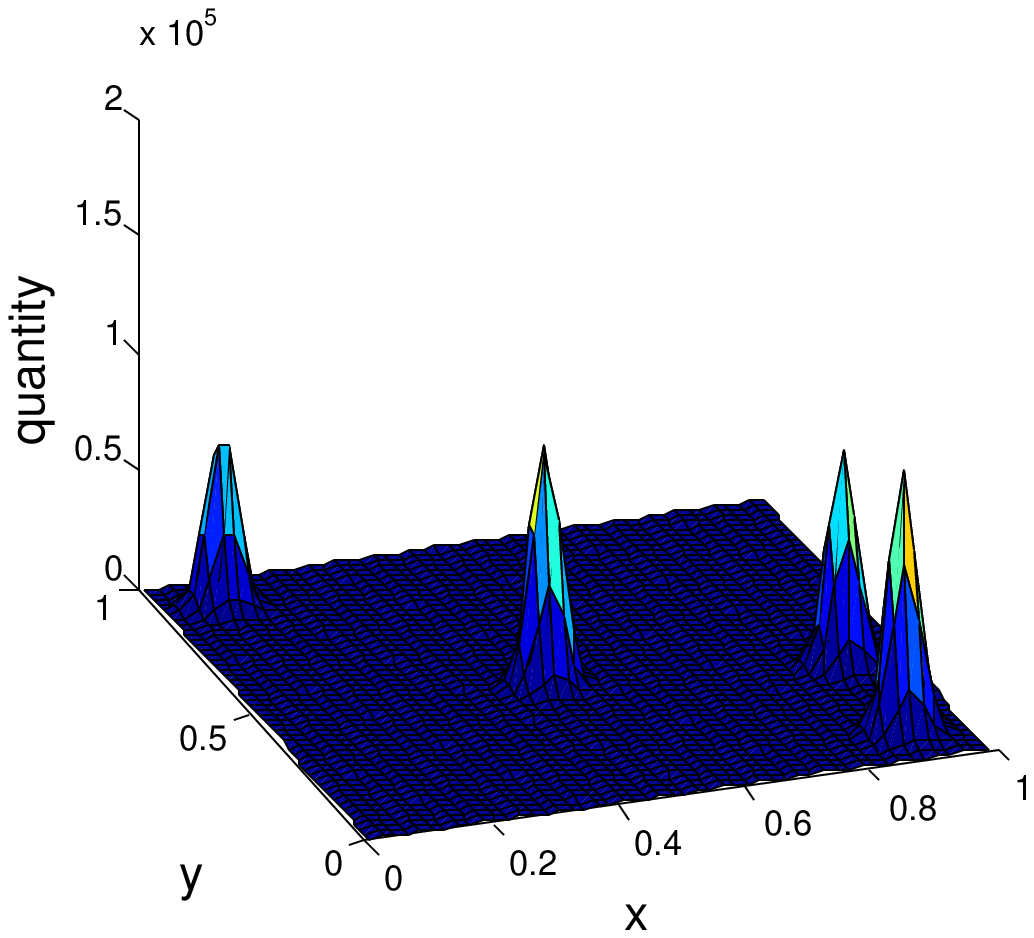}
      } \\
      
      \subfloat[Final Partition, \mi{t=4}]{
        \parbox{0.22\textwidth}{
          \centering
          \includegraphics[width=0.20\textwidth]{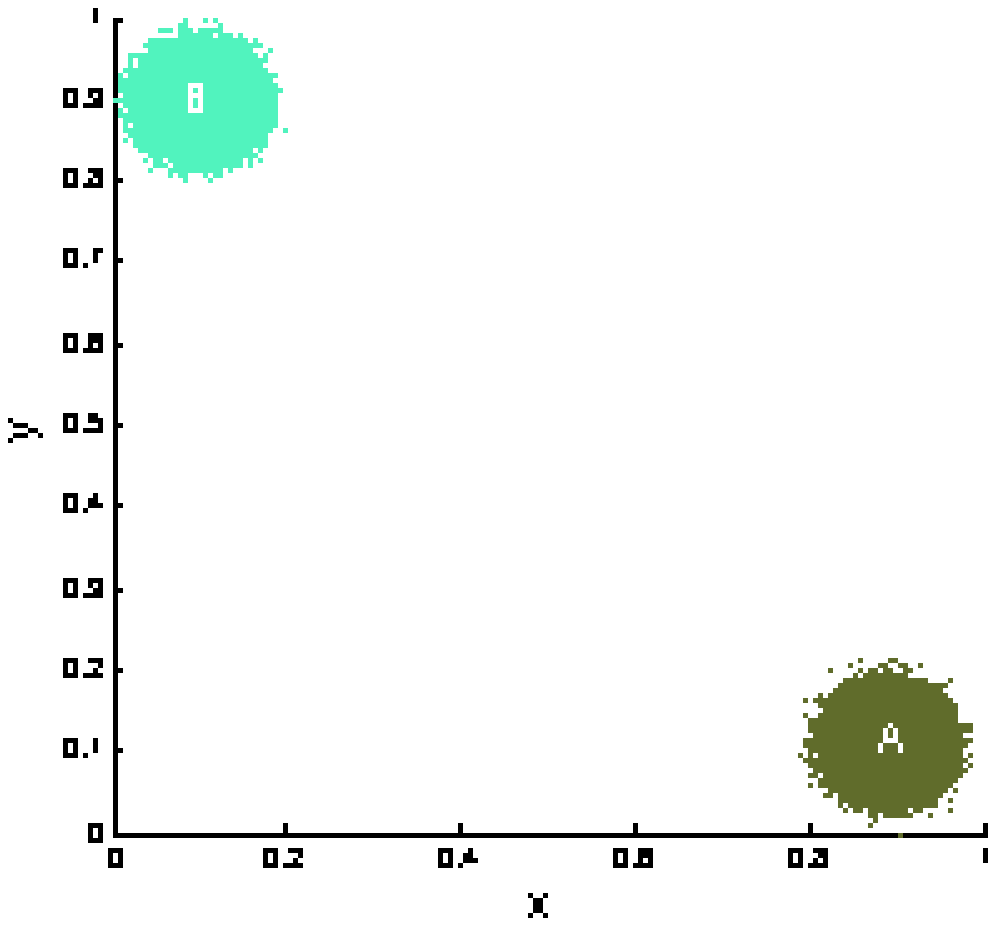}
        }
      }
      \subfloat[Final Partition, \mi{t=8}]{
        \parbox{0.22\textwidth}{
          \centering
          \includegraphics[width=0.20\textwidth]{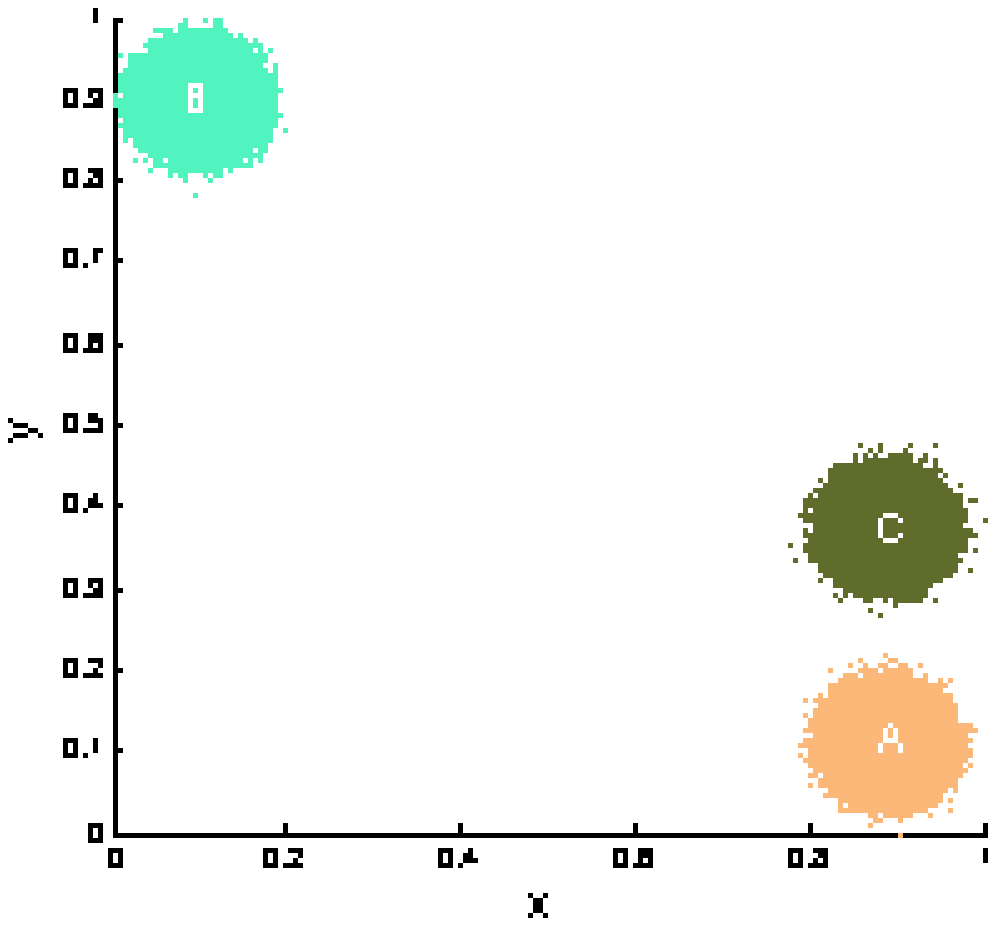}
        }
      }
      \subfloat[Final Partition, \mi{t=12}]{
        \parbox{0.22\textwidth}{
          \centering
          \includegraphics[width=0.20\textwidth]{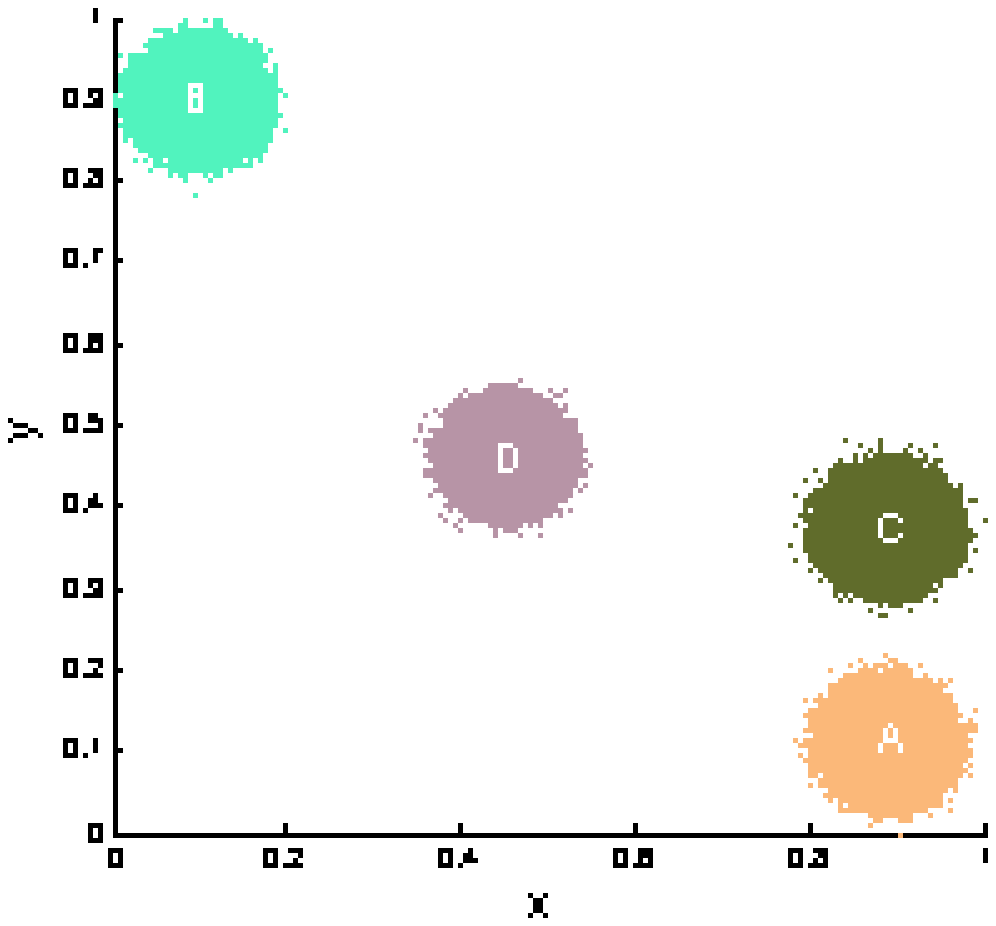}
        }
      }
      \subfloat[Final Partition, \mi{t=16}]{
        \parbox{0.22\textwidth}{
          \centering
          \includegraphics[width=0.20\textwidth]{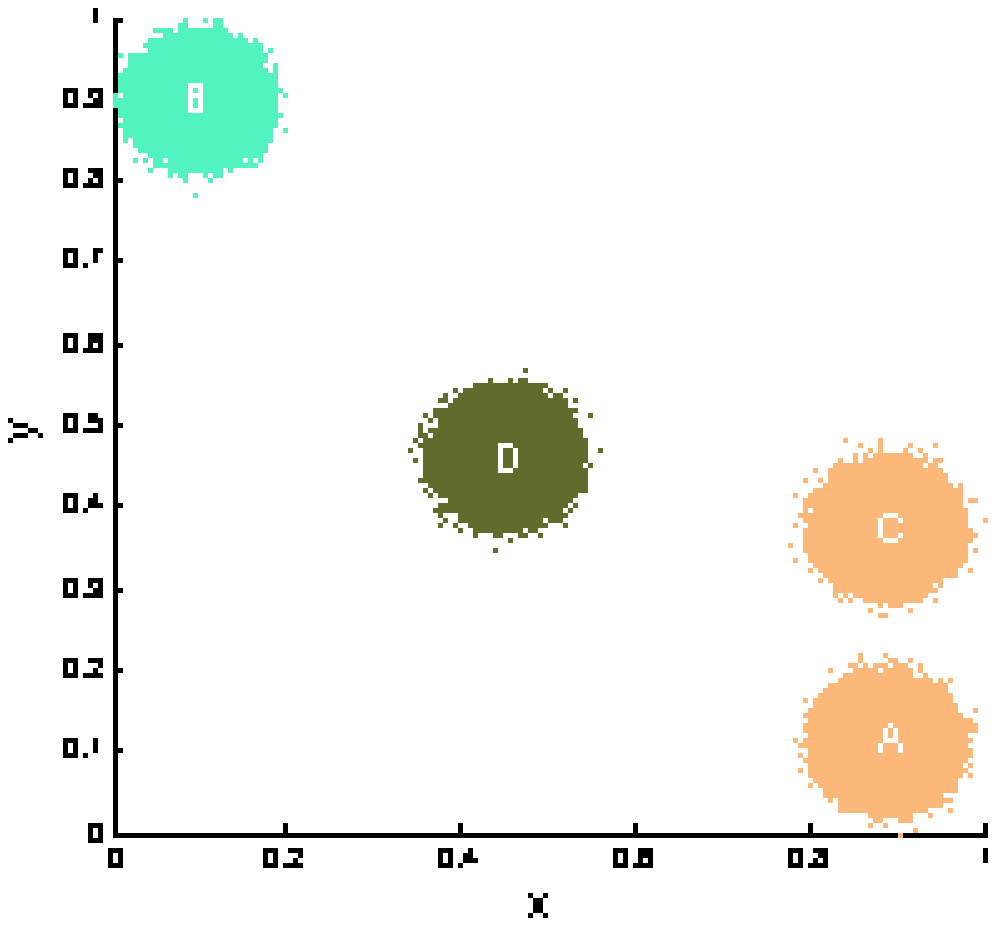}
        }
      }
      \caption{Snapshot of data set and cluster configuration at several moments in time.}
      \label{fig:prkmtvSnapshots}
  \end{center}
\end{figure*}

Figure~\subref*{fig:2d_time_varying_clusters} shows that the number of clusters adapts to accommodate changes in the underlying data set.  The data from subsets A and B are well-separated and the algorithm forms two clusters at first.  As data from subset C starts to accumulate, the algorithm adds a cluster.  Similarly, a fourth cluster is added when data from subset D arrives.  Eventually subsets A and C are small, compared to subsets B and D, and are sufficiently close to be joined into a single cluster.  The \mi{k=3} solution persists until the end of the experiment.

\begin{figure}[htpb]
    \begin{center}
        \includegraphics[width=0.4\textwidth]{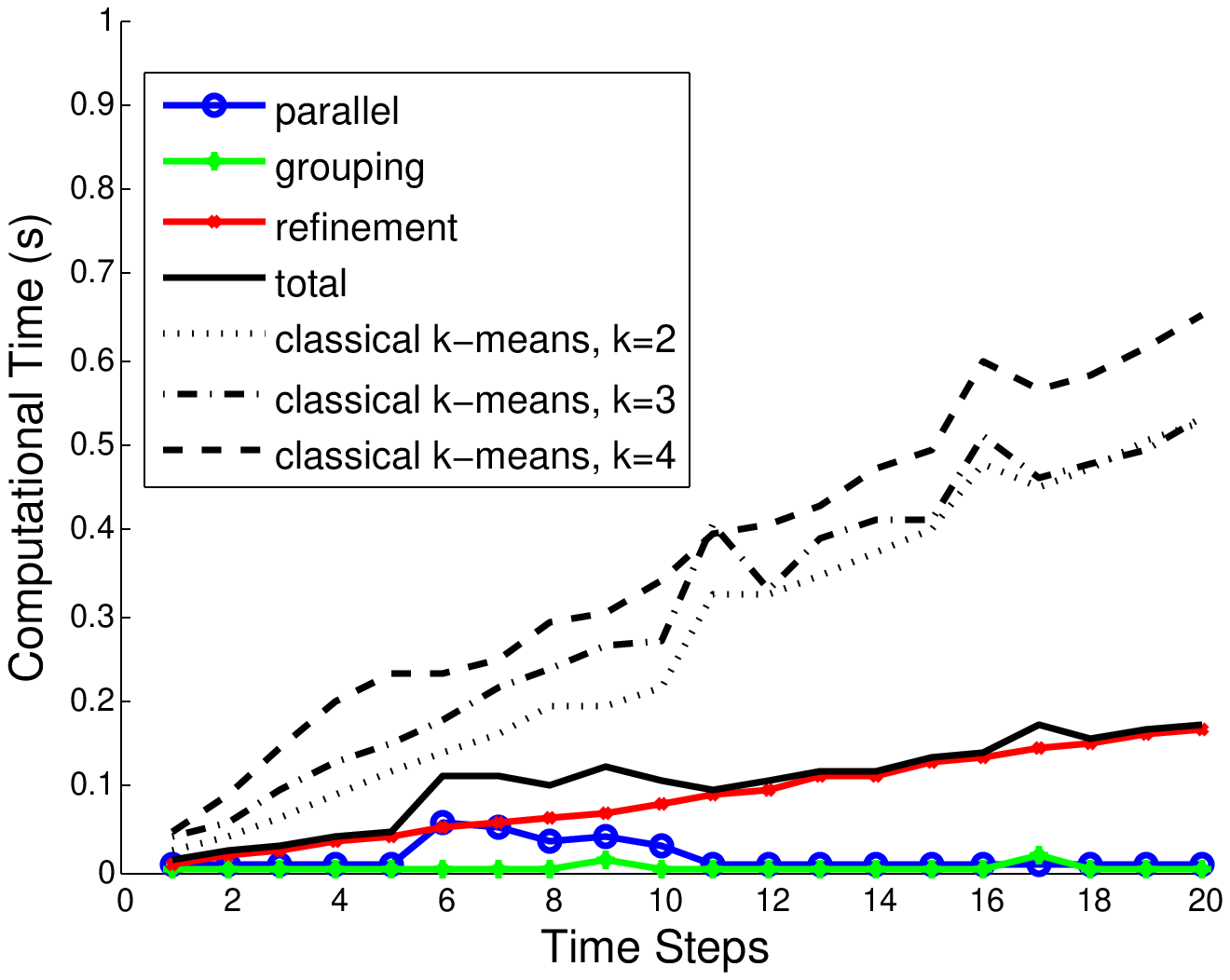}
        \caption{Computational time for clustering time-dependent data set using the proposed algorithm with 16 threads.  For comparison, computational time required to cluster the data at each time step using \mi{k}-means is shown for \mi{k=2,3,4}.  The time-dependent PAC algorithm demonstrates better scaling as the size of the aggregate data increases.}
        \label{fig:2d_time_varying_hpc}
    \end{center}
\end{figure}

The primary benefit of the proposed algorithm for clustering time-dependent data is the ability to dynamically update the number of clusters at each time step.  A secondary benefit is computational efficiency.  When methods such as \mi{k}-means are used to cluster the time-dependent data, it is necessary to recompute the classical \mi{k}-means partition from scratch at each time increment.  For \mi{k=2,3,4}, the computational cost of this approach increases quickly as the data set grows in size.  The cost also increases for the PAC algorithm in Figure~\ref{fig:2d_time_varying_hpc} as the data set grows in size, but more slowly.  The cost of the refinement step increases approximately linearly, but more slowly than the cost of classical \mi{k}-means.  In this experiment, \mi{\lambda_c} is sufficiently small to prevent under-segmentation.  The cost of the parallel step is low and roughly constant as the number of data being processed by this step is constant at each time instant.  The cost of the grouping step is also very small at each time step.  With the selected parameters and 16 parallel threads, the time-dependent parallel adaptive clustering algorithm is faster than the naive application of classical \mi{k}-means, with the added benefit of dynamically adapting the number of clusters.

\section{Conclusion} \label{sec:conclusion}

This work presents the parallel adaptive clustering (PAC) algorithm as a new method for centroid-based clustering that dynamically selects the number of clusters and leverages parallel processing to increase efficiency.  Mathematical analysis of the algorithm gives clear insight into how the regularization parameters affect cluster properties, the basis of the regularized set \mi{k}-means, the stability of the pre-refinement cluster configuration, and a computationally efficient implementation of the refinement procedure.  The PAC algorithm is computationally efficient for multi-core computing architecture, and is immediately applicable to a wide range of data, including data streams.  

\bibliographystyle{plain}
\bibliography{refs}

\begin{thebibliography}{10}

\bibitem{arlia2001experiments}
Domenica Arlia and Massimo Coppola.
\newblock Experiments in parallel clustering with dbscan.
\newblock In {\em European Conference on Parallel Processing}, pages 326--331.
  Springer, 2001.

\bibitem{arthur2007k}
David Arthur and Sergei Vassilvitskii.
\newblock k-means++: The advantages of careful seeding.
\newblock In {\em Proceedings of the eighteenth annual ACM-SIAM symposium on
  Discrete algorithms}, pages 1027--1035. Society for Industrial and Applied
  Mathematics, 2007.

\bibitem{banfield1993model}
Jeffrey~D Banfield and Adrian~E Raftery.
\newblock Model-based gaussian and non-gaussian clustering.
\newblock {\em Biometrics}, pages 803--821, 1993.

\bibitem{ding2015yinyang}
Yufei Ding, Yue Zhao, Xipeng Shen, Madanlal Musuvathi, and Todd Mytkowicz.
\newblock Yinyang k-means: A drop-in replacement of the classic k-means with
  consistent speedup.
\newblock In {\em Proceedings of the 32nd International Conference on Machine
  Learning (ICML-15)}, pages 579--587, 2015.

\bibitem{dunn1974well}
Joseph~C Dunn.
\newblock Well-separated clusters and optimal fuzzy partitions.
\newblock {\em Journal of cybernetics}, 4(1):95--104, 1974.

\bibitem{ester1996density}
Martin Ester, Hans-Peter Kriegel, J{\"o}rg Sander, Xiaowei Xu, et~al.
\newblock A density-based algorithm for discovering clusters in large spatial
  databases with noise.
\newblock In {\em Kdd}, volume~96, pages 226--231, 1996.

\bibitem{everitt2011cluster}
B.S. Everitt, S.~Landau, M.~Leese, and D.~Stahl.
\newblock {\em Cluster Analysis}.
\newblock Wiley Series in Probability and Statistics. Wiley, 2011.

\bibitem{fraley1998many}
Chris Fraley and Adrian~E Raftery.
\newblock How many clusters? which clustering method? answers via model-based
  cluster analysis.
\newblock {\em The computer journal}, 41(8):578--588, 1998.

\bibitem{garg2006pbirch}
Ashwani Garg, Ashish Mangla, Neelima Gupta, and Vasudha Bhatnagar.
\newblock Pbirch: A scalable parallel clustering algorithm for incremental
  data.
\newblock In {\em 2006 10th International Database Engineering and Applications
  Symposium (IDEAS'06)}, pages 315--316. IEEE, 2006.

\bibitem{guha2003clustering}
Sudipto Guha, Adam Meyerson, Nina Mishra, Rajeev Motwani, and Liadan
  O'Callaghan.
\newblock Clustering data streams: Theory and practice.
\newblock {\em IEEE transactions on knowledge and data engineering},
  15(3):515--528, 2003.

\bibitem{hartigan1979algorithm}
John~A Hartigan and Manchek~A Wong.
\newblock Algorithm {AS} 136: A k-means clustering algorithm.
\newblock {\em Journal of the Royal Statistical Society. Series C (Applied
  Statistics)}, 28(1):100--108, 1979.

\bibitem{jain1988algorithms}
Anil~K Jain and Richard~C Dubes.
\newblock {\em Algorithms for clustering data}.
\newblock Prentice-Hall, Inc., 1988.

\bibitem{kang2011regularized}
Sung~Ha Kang, Berta Sandberg, and Andy~M Yip.
\newblock A regularized k-means and multiphase scale segmentation.
\newblock {\em Inverse Problems and Imaging (IPI)}, 5(2):407--429, 2011.

\bibitem{kantabutra2000parallel}
Sanpawat Kantabutra and Alva~L Couch.
\newblock Parallel k-means clustering algorithm on nows.
\newblock {\em NECTEC Technical journal}, 1(6):243--247, 2000.

\bibitem{kaufman2009finding}
Leonard Kaufman and Peter~J Rousseeuw.
\newblock {\em Finding groups in data: an introduction to cluster analysis},
  volume 344.
\newblock John Wiley \& Sons, 2009.

\bibitem{li1989parallel}
Xiaobo Li and Zhixi Fang.
\newblock Parallel clustering algorithms.
\newblock {\em Parallel Computing}, 11(3):275--290, 1989.

\bibitem{li2013speeding}
You Li, Kaiyong Zhao, Xiaowen Chu, and Jiming Liu.
\newblock Speeding up k-means algorithm by gpus.
\newblock {\em Journal of Computer and System Sciences}, 79(2):216--229, 2013.

\bibitem{lloyd1982least}
Stuart Lloyd.
\newblock Least squares quantization in pcm.
\newblock {\em IEEE transactions on information theory}, 28(2):129--137, 1982.

\bibitem{macqueen1967some}
James MacQueen.
\newblock Some methods for classification and analysis of multivariate
  observations.
\newblock In {\em Proceedings of the fifth Berkeley symposium on mathematical
  statistics and probability}, number~14 in 1, pages 281--297. Oakland, CA,
  USA., 1967.

\bibitem{olman2008parallel}
Victor Olman, Fenglou Mao, Hongwei Wu, and Ying Xu.
\newblock Parallel clustering algorithm for large data sets with applications
  in bioinformatics.
\newblock {\em IEEE/ACM Transactions on Computational Biology and
  Bioinformatics}, 6(2):344--352, 2008.

\bibitem{olson1995parallel}
Clark~F Olson.
\newblock Parallel algorithms for hierarchical clustering.
\newblock {\em Parallel computing}, 21(8):1313--1325, 1995.

\bibitem{pena1999empirical}
Jos{\'e}~M Pena, Jose~Antonio Lozano, and Pedro Larranaga.
\newblock An empirical comparison of four initialization methods for the
  k-means algorithm.
\newblock {\em Pattern recognition letters}, 20(10):1027--1040, 1999.

\bibitem{rousseeuw1987silhouettes}
Peter~J Rousseeuw.
\newblock Silhouettes: a graphical aid to the interpretation and validation of
  cluster analysis.
\newblock {\em Journal of computational and applied mathematics}, 20:53--65,
  1987.

\bibitem{sandberg2010unsupervised}
Berta Sandberg, Sung~Ha Kang, and Tony~F Chan.
\newblock Unsupervised multiphase segmentation: A phase balancing model.
\newblock {\em IEEE transactions on image processing}, 19(1):119--130, 2010.

\bibitem{sparks1973algorithm}
DN~Sparks.
\newblock Algorithm as 58: Euclidean cluster analysis.
\newblock {\em Journal of the Royal Statistical Society. Series C (Applied
  Statistics)}, 22(1):126--130, 1973.

\bibitem{stoffel1999parallel}
Kilian Stoffel and Abdelkader Belkoniene.
\newblock Parallel k/h-means clustering for large data sets.
\newblock In {\em European Conference on Parallel Processing}, pages
  1451--1454. Springer, 1999.

\bibitem{zhang2013parallel}
Jing Zhang, Gongqing Wu, Xuegang Hu, Shiying Li, and Shuilong Hao.
\newblock A parallel clustering algorithm with mpi-mkmeans.
\newblock {\em J Comput}, 8(1):10--17, 2013.

\bibitem{zhang1996birch}
Tian Zhang, Raghu Ramakrishnan, and Miron Livny.
\newblock Birch: an efficient data clustering method for very large databases.
\newblock In {\em ACM Sigmod Record}, volume~25, pages 103--114. ACM, 1996.

\bibitem{zhao2009parallel}
Weizhong Zhao, Huifang Ma, and Qing He.
\newblock Parallel k-means clustering based on mapreduce.
\newblock In {\em IEEE International Conference on Cloud Computing}, pages
  674--679. Springer, 2009.

\end{thebibliography}

\end{document}